\documentclass[journal]{IEEEtran}

\usepackage{microtype}
\usepackage{graphicx}
\usepackage{subfigure}
\usepackage{booktabs} % for professional tables
\usepackage{wrapfig}

\usepackage{hyperref}
\usepackage[numbers]{natbib}  % Use numbers instead of author-year for IEEE

\usepackage{amsmath}
\usepackage{amssymb}
\usepackage{mathtools}
\usepackage{amsthm}
\newcommand{\ra}[1]{\renewcommand{\arraystretch}{#1}}
\usepackage{pifont}% http://ctan.org/pkg/pifont
\newcommand{\cmark}{\ding{51}}%
\newcommand{\xmark}{\ding{55}}%
\usepackage{colortbl} % for coloring cells
\definecolor{darkgreen}{rgb}{0.0, 0.5, 0.0}
\definecolor{lightgray}{gray}{0.9}
\usepackage{makecell} % for thick lines
\usepackage{xcolor}
\usepackage[capitalize,noabbrev]{cleveref}
\newcolumntype{C}[1]{>{\centering\let\newline\\\arraybackslash\hspace{0pt}}m{#1}}

\theoremstyle{plain}

\theoremstyle{definition}

\theoremstyle{remark}

\usepackage[textsize=tiny]{todonotes}
\title{MobilityGPT: Enhanced Human Mobility \\ Modeling with a GPT model}

\begin{document}

\author{Ammar~Haydari, Dongjie~Chen,  Zhengfeng~Lai, Michael~Zhang, ~Chen-Nee~Chuah,~\IEEEmembership{Fellow,~IEEE}

\thanks{Ammar Haydari, Dongjie, Chen and Chen-Nee Chuah are with the Department
of Electrical and Computer Engineering, University of California, Davis,
CA, 95616 USA (e-mail:ahaydari@ucdavis.edu; cdjchen@ucdavis.edu; chuah@ucdavis.edu).}

\thanks{Zhengfeng Lai is with Apple Inc. Cupertino, CA, 95014 USA (e-mail: lzhengfeng@ucdavis.edu).}

\thanks{Michael Zhang is with the Department of Civil and Environmental Engineering, University of California, Davis, CA 95616 USA (e-mail: hmzhang@ucdavis.edu).}
}

\maketitle

\begin{abstract}

Generative models have shown promising results in capturing human mobility characteristics and generating synthetic trajectories. However, it remains challenging to ensure that the generated geospatial mobility data is semantically realistic, including consistent location sequences, and reflects real-world characteristics, such as constraining on geospatial limits. We reformat human mobility modeling as an autoregressive generation task to address these issues, leveraging the Generative Pre-trained Transformer (GPT) architecture. To ensure its controllable generation to alleviate the above challenges, we propose a geospatially-aware generative model, MobilityGPT. We propose a gravity-based sampling method to train a transformer for semantic sequence similarity. Then, we constrained the training process via a road connectivity matrix that provides the connectivity of sequences in trajectory generation, thereby keeping generated trajectories in geospatial limits. Lastly, we proposed to construct a preference dataset for fine-tuning MobilityGPT via Reinforcement Learning from Trajectory Feedback (RLTF) mechanism, which minimizes
the travel distance between training and the synthetically generated trajectories. Experiments on real-world datasets demonstrate MobilityGPT's superior performance over state-of-the-art methods in generating high-quality mobility trajectories that are closest to real data in terms of origin-destination similarity, trip length, travel radius, link, and gravity distributions. We release the source code and reference links to datasets at \url{https://github.com/ammarhydr/MobilityGPT}. 
\end{abstract}


\section{Introduction}

The widespread integration of location-based services and smart GPS devices, such as smartphones and watches, has made continuous monitoring of human mobility both desirable and feasible. These technologies capture diverse and detailed human movement information, with mobility trajectories representing the finest granularity of individual-level mobility characteristics. Such trajectories are crucial in various applications, including mobility modeling, commercial business analysis, and disease spread control \citep{uppoor2013generation, wang2021survey, zhou2024unified}. Understanding and modeling these mobility patterns has become increasingly critical as population growth and complex travel behaviors reshape modern transportation systems. The impacts of human mobility extend beyond transportation, influencing urban planning, environmental quality, public health, and energy consumption. Recent global events, particularly the COVID-19 pandemic, have highlighted the dynamic nature of mobility patterns and their profound societal implications \cite{zhang2021effect, hu2021human}. This has intensified the need for sophisticated mobility modeling approaches that can capture, analyze, and predict human movement patterns at various spatial and temporal scales.

Despite increasing demand for human mobility trajectory datasets, numerous challenges hinder their access and distribution \citep{kong2023mobility, song2017multi}. First, these datasets are typically collected by private companies or government agencies, hence presenting privacy concerns due to the potential disclosure of an individual's sensitive lifestyle patterns (e.g., home or office addresses and points of interest). Legal implications, such as those outlined in the California Consumer Privacy Act \cite{bonta2022california} and General Data Protection Regulation \cite{GDPR2016}, highlight the importance of carefully handling spatiotemporal traces.  Second, datasets owned by companies may expose proprietary business models and are often inaccessible for research purposes~\citep{wang2019urban}. Lastly, publicly available datasets often lack diversity or quality, with gaps in data points and intrinsic noise that significantly reduce their utility. These limitations impede the progress of urban planning or transportation research \citep{luca2021survey}. Therefore, it is imperative to establish alternative trajectory data sources that are both high-quality and accessible for research purposes. 

\begin{figure}
    \centering
    \includegraphics[width=0.9\linewidth]{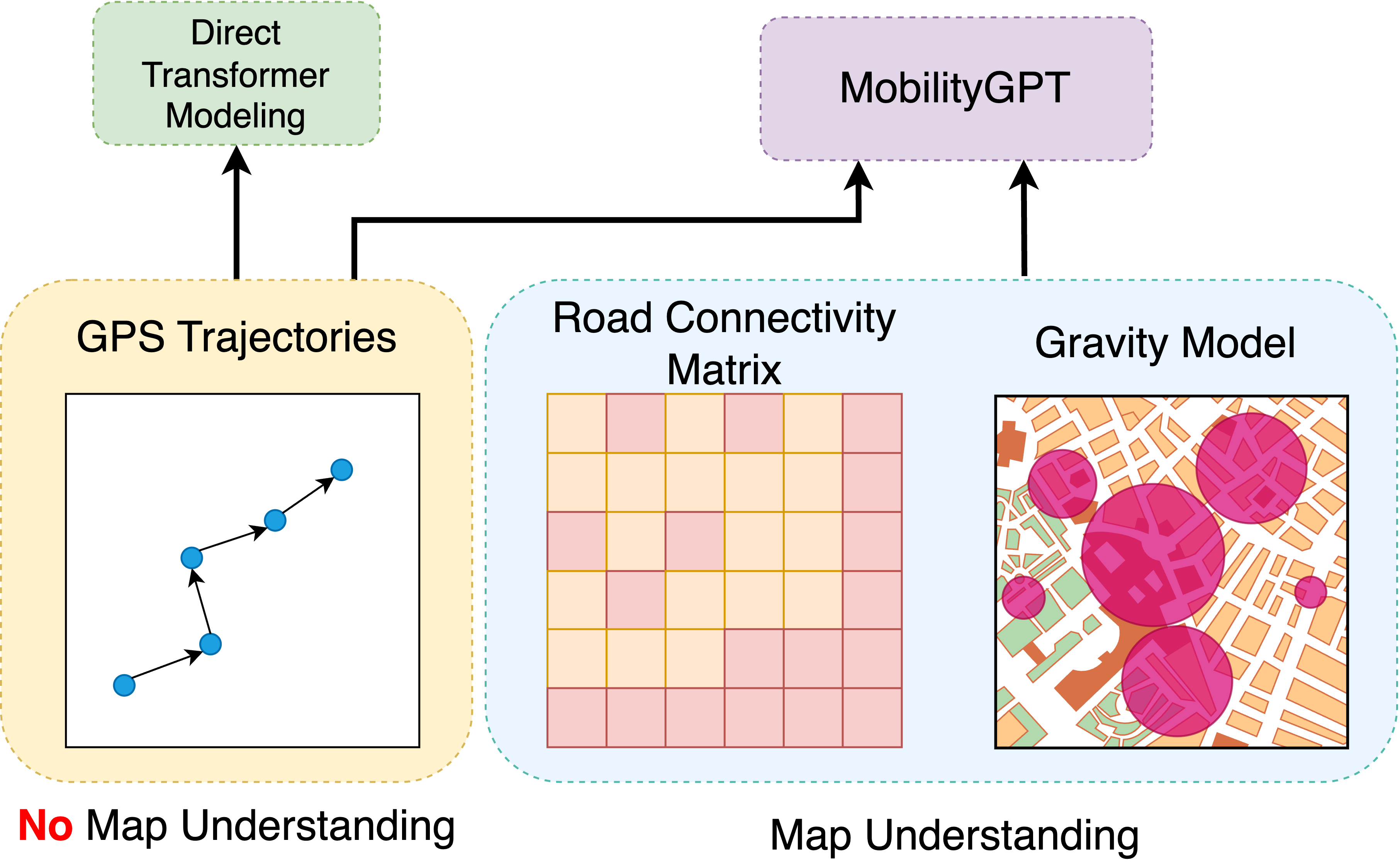}
    \caption{Generative models that do not incorporate mobility characteristics struggle to capture human mobility patterns accurately.}
    \label{fig:spotlight}
\end{figure}

Current generative models designed for human mobility exhibit multiple drawbacks: they fail to capture sequential mobility characteristics, often lack the continuity of generated trajectories, and do not consistently adhere to geospatial constraints. Approaches based on the Generative Adversarial Network (GAN) \citep{goodfellow2020generative} and the Variational Autoencoder (VAE) \citep{kingma2013auto} structure data in a tabular format, which unfortunately fails to preserve the inherent correlations between locations. Models like Long-Short-Term Memory (LSTM) and sequential GAN \citep{yu2017seqgan} struggle to maintain smooth continuity of locations and capture realistic human mobility characteristics. A further limitation of these methods lies in the elevated mismatches observed in geospatial location sequences. In short, a successful generative model for trajectory datasets should precisely grasp the intrinsic spatial-temporal behaviors.

\textcolor{black}{GPT (Generative Pre-trained Transformer) is a language model architecture that learns to generate coherent sequences by predicting the next token in a sequence, based on the context of previous tokens. It is trained in an autoregressive manner and excels at modeling sequential dependencies. This approach offers a robust foundation for sequence generation tasks and can be refined through fine-tuning to cater to distinct objectives.} 

Trajectories and sentences share several similarities that make GPTs a promising approach for trajectory generation. Both consist of ordered sets of elements chosen from finite pools (road links and words, respectively). They exhibit semantic or spatiotemporal relationships, adhering to distinct rule systems such as language rules for sentences and geographical constraints for trajectories. Due to these parallels, the techniques developed for natural language processing can be adapted to model and generate realistic trajectories \citep{musleh2022let}. In this work, we adapt the GPT architecture to the domain of human mobility by treating trajectories as sequences of spatial tokens (e.g., road links). We found that the self-attention mechanism captures intricate dependencies in human mobility sequences. This positions such models as powerful tools for overcoming challenges in human trajectory generation. Besides, the gravity model holds significance in human mobility modeling as it provides a structured framework for estimating and understanding the flow of interactions between different locations \citep{zipf1946p}. Using gravity as part of human mobility modeling could bring further advantages to synthetic trajectory generation tasks.   

We leverage the power of GPT architecture for synthetic trajectory generation. However, directly training a GPT on mobility data without geospatial insights would generate unrealistic sequences. To generate more realistic synthetic trajectories, we introduced several innovative methods, including (1) gravity-aware sampling, which incorporates gravity modeling of trajectory data into training updates, and (2) the use of road connectivity matrix (RCM) masking to eliminate disconnected location sequences from logits (see Figure \ref{fig:spotlight}). Pretraining GPT models on trajectory data captures the general sequence characteristics in an unstructured manner. We, furthermore, proposed an automated fine-tuning pipeline that improves the trajectory quality by leveraging transportation-specific metrics to evaluate and optimize the generated sequences without relying on human labeling. Our multi-objective framework, MobilityGPT, not only extends the application of GPT models to human mobility modeling but also collectively enhances the model's capacity to capture intricate patterns and adhere to realistic geospatial constraints (e.g., trajectories consist of road segments that are indeed connected). MobilityGPT can generate contextually rich and semantically accurate trajectories that reflect realistic human mobility without requiring prohibitive computational complexity.

\textcolor{black}{\textbf{Contributions:} In summary, this paper addresses concerns like trajectory continuity, adherence to geospatial constraints, and computational complexity in existing models. The core of our contribution is MobilityGPT, a human mobility modeling framework that features geospatially aware pretraining and automated fine-tuning without requiring human feedback. We enhance trajectory generation by implementing gravity-aware sampling that trains our generative model using Origin-Destination (OD) pair-gravity values, while ensuring trajectory continuity through next sequence prediction conditioned on the road network through RCM. To improve trajectory quality, we developed a strategy to construct preference datasets for fine-tuning MobilityGPT via reinforcement learning, which significantly enhances travel length similarity in generated paths. Our comprehensive experimental analysis with real-world datasets demonstrates that these methods produce high-fidelity trajectories that maintain both the statistical properties and semantic characteristics essential to authentic human mobility patterns.}

\textcolor{black}{The remainder of the paper is organized as follows. We present related work in Section \ref{s:related}, establish our preliminary definitions in Section \ref{s:metrics}, and introduce MobilityGPT components in Section \ref{s:MobilityGPT}. After empirical evaluation in Section \ref{s:results}, we discuss limitations in Section \ref{s:discuss}. Finally, Section \ref{s:conclusion} concludes the paper.}

\section{Related Work}
\label{s:related}

\textcolor{black}{Human mobility simulation presents a significant challenge in mobility data mining due to complex geospatial and temporal correlations in movement sequences. Existing methods aim to generate realistic trajectories that reflect human behavior. Many focus on movement prediction \cite{ying2011semantic, Qingyue2025Universal} or trajectory reconstruction \cite{jiang2023continuous}. However, they often miss dynamic mobility patterns. They also struggle to generalize across large-scale datasets, limiting their use in comprehensive mobility modeling.}

\textcolor{black}{The methods for synthetic mobility trajectory generation can be broadly categorized into knowledge-driven \textit{model-based} and data-driven \textit{model-free} approaches. The latter predominantly leveraging machine learning techniques. \textit{Model-based} methods rely on theoretical frameworks and domain-specific hypotheses, such as gravity models and spatial interaction theory, to simulate mobility patterns. Examples include microscopic traffic simulators like SUMO \cite{lopez2018microscopic} and VISSIM \cite{fellendorf2010microscopic}. These methods, however, often struggle to capture the complexity and variability of real-world human mobility, particularly at finer granularities.}

\textcolor{black}{There exists another important distinction in the trajectory literature between point-based and point-of-interest-based generative models. Point-based location sequences are dense and contain richer human movement information at high granularity. Recently, TrajGPT \cite{hsu2024trajgpt} proposed a visitation-based sequence generation model that generates sparse location sequences based on visited locations. However, this method falls short in dense sequence prediction tasks. MobilityGPT takes a fundamentally different approach from these works, addressing the limitations of sparse sequence models by focusing on continuous, high-resolution trajectory generation that captures the complete dynamics of human movement patterns.}

\textcolor{black}{Early data-driven \textit{model-free} approaches, employing techniques such as gravity models, decision trees, and Markovian methods such as Mobility Markov Chains (MMC) \citep{pappalardo2016human, jiang2016timegeo, gambs2012next}, face challenges in capturing sequential transitions between locations. In contrast, generative machine learning models have gained traction for their ability to learn from data without external input parameters \citep{baratchi2014hierarchical}. Such generative models have been explored for human mobility modeling using various data representations, including grids in tabular format \citep{ouyang2018non}, image-like trajectory modeling \citep{cao2021generating}, and sequential grid format \citep{feng2020learning}. However, these models exhibit limitations in generating trajectories that accurately capture geospatial complexity.}

\textcolor{black}{One promising generative model direction relies on GPS sampling. Due to mismatches, LSTM-based trajectory generation models cannot provide continuous motion of human mobility in diverse geospatial areas \citep{rao2020lstm}. A similar continuous GPS trajectory generation method with a U-Net neural network using Diffusion models is proposed in \citep{zhu2024difftraj}. Sampling GPS locations without geospatial constraints diminishes the quality of synthetic trajectories. They rarely incorporate geospatial constraints and road network topologies into their training pipeline, resulting in physically impossible trajectories that ignore natural barriers and transportation infrastructure limitations when applied to new environments.}

\textcolor{black}{Another generative direction in ML for mobility trajectories employs two-stage training methods \citep{wang2021large, jiang2023continuous, long2023practical}. A recent GAN-based model that relies on road links with the $A^{*}$ path-finding algorithm between regions using two-stage GANs has shown promise in identifying diverse paths for specified OD pairs \citep{jiang2023continuous}. It falls short in generating truly synthetic trajectories, as the generation process requires initiating with an OD pair from testing trajectories. Another two-stage approach generates mobility trajectories using dual VAEs conditioned on user-level mobility characteristics learned in the second VAE model \citep{long2023practical}. This approach is limited by its requirement for learning user-driver behaviors across multiple trips, necessitating substantial data per individual. In contrast, our MobilityGPT model learns generalist human mobility patterns without requiring personalization, making it more adaptable and efficient with available data.}

\textcolor{black}{Finally, sequential GAN methods have also been employed for mobility modeling with different variants \citep{yu2017seqgan, feng2020learning}. These earlier models generate data in sequences but struggle to capture geospatial complexity and add significant computational complexity, as discussed later in our ablation studies \ref{tab:comp_demand}. Unlike previous methods, MobilityGPT leverages the autoregressive GPT model for trajectory generation. This enables a comprehensive exploration of spatial and temporal distributions with high-quality outputs by incorporating road links into sequence modeling.}

The GPT models,  developed by \citep{radford2018improving}, transformed the field of Natural Language Processing (NLP) with the transformer's capabilities. GPT, with its pre-training on diverse language data, demonstrated remarkable proficiency in understanding and generating coherent sequences to support various NLP tasks. 
The transformer-based GPT concept has been applied to diverse applications, from vision \citep{chen2020generative}, music \citep{banar2022systematic} to network data \citep{kholgh2023pac}. In addition, researchers have made great progress on various fine-tuning models for enhancing the GPT model capabilities \citep{zhang2019ernie, brown2020language, lambert2022illustrating}. However, these fine-tuning models mainly rely on human feedback that evaluates the quality of the generated sequences. This study proposes a geospatially-aware GPT model, MobilityGPT, with an innovative fine-tuning approach for generating synthetic mobility trajectories.

\section{Preliminaries}
\label{s:metrics}

\subsection{Discretization with Map-Matching}

Let $D(V,E)$ represent the road network as a weighted digraph, where the set of nodes $V$ corresponds to the intersection of the road, the set of edges $E$ to the roads and the weights representing the metrics of the link, such as the length of the link or traffic volume. There are two sets of trajectories: GPS trajectories and link trajectories. 

{\bf 1) GPS trajectories}: A sequence of GPS coordinates with $l$ number of samples $ \mathbf{x} \in \mathcal{G} = \{\mathbf{x}_{1}, \mathbf{x}_{2}, ..., \mathbf{x}_{l}\}$ forms a GPS trajectory that reflects the continuous motion of the object. 

{\bf 2) Link trajectories}: Given the $s$ number of vehicles on the road network, each vehicle travels between origins and destinations using an ordered link path generating a user travel path known as a \textit{micro-graph} $\Phi \in D$. Every link trajectory has $n$ number of links $\phi \in \Phi = \{\phi_{1}, \phi_{2}, ..., \phi_{n}\}$ and $\Phi \subset{E}$. The set of trajectories is the corpus of all link trajectories with $s$ users $\Phi \in \Lambda=\{\Phi_{1}, \Phi_{2}, ..., \Phi_{s}\}$.

{\bf 3) Map-matching}: GPS coordinates provide an approximation of a device's location by utilizing satellite broadcast information. Nonetheless, these coordinates may not consistently reflect the precise travel path, as they are susceptible to intrinsic and environmental errors, including satellite geometry, signal blockage, or tree cover \citep{berber2012comparison}. Consequently, the recorded GPS locations might not align with a specific link on the road network. To address this, map-matching is employed to generate an ordered set of road network links that accurately describes the user's trajectory, taking into account the road network $D(V,E)$ and GPS points \citep{quddus2007current}.

Addressing location sequences in the continuous domain poses challenges. Consequently, conventional human mobility modeling methods often resort to the discretized grid domain. However, relying on a grid-like discrete representation does not eliminate the risk of geospatial mismatching. For instance, random sampling from a sparse road network grid may lead to nonsensical locations in generated trajectories. In response to this limitation, our MobilityGPT model adopts a strategy of discretizing locations into road segments, enhancing the quality of generated trajectories.

\subsection{Problem Definition}

In an urban setting, a generative model is tasked with learning the intrinsic mobility characteristics from a provided set of link trajectories, $\Phi \in \Lambda = \{\Phi_{1}, \Phi_{2}, ..., \Phi_{s}\}$. The objective is to model human mobility using a sequential learning approach and generate synthetic trajectories that resemble training data across key utility metrics.

The sequential nature of mobility trajectories encapsulates human routing behavior, where the successive road segments influence subsequent link choice. The generative model aims to learn the conditional probability distribution $P(\phi_{k+1}|\phi_k, \phi_{k-1}, ..., \phi_{k-m})$, capturing the sequential dependencies that characterize human mobility patterns. The generation process involves producing synthetic trajectories that maintain both network connectivity constraints and realistic routing behaviors observed in the training data. 

\section{Mobility Modeling with Generative Transformers}
\label{s:MobilityGPT}

\begin{figure*}[t]
    \centering
    \includegraphics[width=1\linewidth]{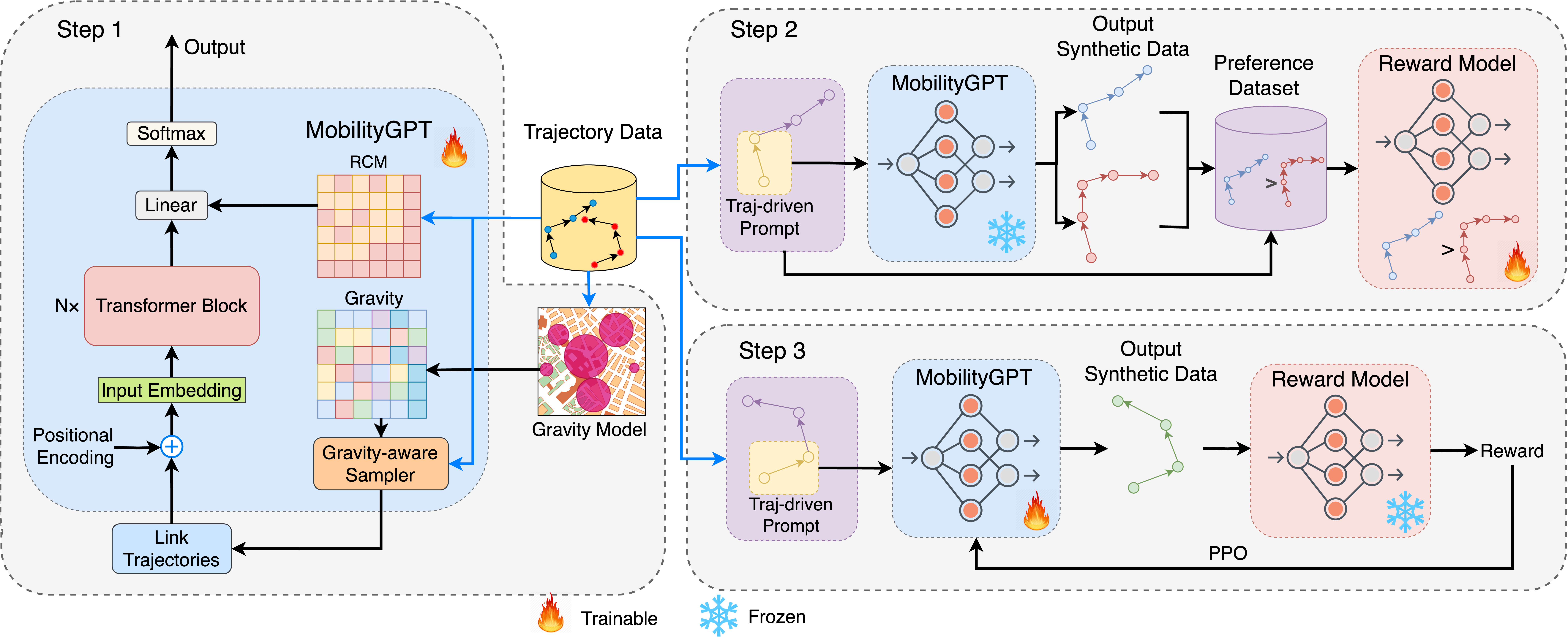}
    \vskip -0.1in
    \caption{Overview. The proposed MobilityGPT framework employs Generative Pretrained Transformer (GPT) architectures and self-adaptive reinforcement learning (RL) fine-tuning methods within a human mobility-aware training pipeline for synthetic trajectory generation. The process involves three steps: 1) Pretraining the GPT model with a gravity model and road connectivity matrix, 2) Constructing a trajectory-driven reward model, and 3) Fine-tuning the MobilityGPT pre-trained model using RL policy optimization methods with the trained reward model providing valuable feedback. }
    \label{fig:MobilityGPT_overall}
    %\vskip -0.2in
\end{figure*}

This section describes the components of the proposed MobilityGPT for generating synthetic high-fidelity trajectories. MobilityGPT achieves this through a multi-objective training pipeline. Figure~\ref{fig:MobilityGPT_overall} illustrates the framework of our MobilityGPT mechanism, which can be divided into two parts: pretraining for learning sequence characteristics and fine-tuning for enhancing the quality of generated trajectories. The pretraining stage (Step 1 in Figure~\ref{fig:MobilityGPT_overall}) learns the mobility modeling on the road-link sequences using the gravity of training data and RCM. The fine-tuning stage (Steps 2 and 3 in Figure~\ref{fig:MobilityGPT_overall}) aims to improve the similarity of synthetic trajectories to the training trajectory samples in terms of travel length.

\subsection{Generative Transformers}

Mobility trajectories are the sequential records of locations or movements over time, capturing the spatiotemporal patterns of an entity's mobility. Their characteristics are crucial in human mobility modeling, providing insights into individual or collective movement behaviors. 

Recent breakthroughs in NLP have proven that transformer architectures are remarkably effective in processing word sequences \citep{zhao2023survey}. Trajectory sequences, representing the spatiotemporal movements of individuals or objects, share similarities with sentences across four key dimensions: sequential dependencies, spatial relationships, contextual embeddings, and variable-length sequences. First, trajectory sequences exhibit sequential dependencies similar to sentences, where the order of locations matters. Additionally, trajectories involve spatial relationships between locations, just as words in a sentence convey semantic relationships. Furthermore, trajectory sequences, like sentences, can have variable lengths. Finally, trajectories benefit from contextual embeddings that consider the entire sequence, similar to the contextual understanding of words in sentences.

Given these strong similarities with trajectories and word sequences, two key elements of transformers, the self-attention mechanism, and autoregressive generation, empower MobilityGPT to handle sequential data efficiently. The self-attention mechanism captures the long-range dependencies and weighs the importance of each location in the context of the entire trajectory, enabling the modeling of sequential dependencies. Besides, autoregressive sequence generation involves predicting one element at a time based on the context of preceding elements, contributing to the model's ability to generate coherent and contextually relevant trajectories.

The self-attention mechanism computes a weighted sum of values $V$ based on attention scores $A$ assigned to each element in the input sequence:
\begin{equation}
\text{Attention}(Q, K, V) = \text{softmax}\left(\frac{QK^T}{\sqrt{d_k}}\right) \cdot V
\end{equation}
where $Q$, $K$, and $V$ represent the query, key, and value matrices. The softmax operation normalizes the attention scores, and $d_k$ is the dimensionality of the key vectors. The weighted sum is then used as the output of the self-attention mechanism.

In an autoregressive nature of transformers, the probability of predicting the next element $y_t$ in the sequence given the context $Y_{<t}$ is computed using the chain rule of probability:
\begin{equation}
    P(y_t | Y_{<t}) = P(y_1) \cdot P(y_2 | y_1) \cdot \ldots \cdot P(y_t | y_{t-1})
\end{equation}

By formulating human mobility trajectories as a sequence generation task with road links from $D(V, E)$, transformer-based models can be applied to generate realistic and contextually relevant synthetic trajectories given the real trajectories. The tokenization of road links, combined with the self-attention mechanism, enables the model to capture spatial dependencies and generate coherent and meaningful sequences of road links. Next, we explain our unique tokenizer modeling.  

\subsection{Tokenizer Modeling}

Tokenization, the breakdown of sequences into smaller units or tokens, is a key step in preparing raw data for transformer models. The influence of tokenization extends to trajectory sequences, where tokens represent locations or spatiotemporal points.

In MobilityGPT, tokens symbolize locations in terms of road links, and the contextual embeddings derived from tokenization contribute to generating embeddings that consider the holistic context of the entire trajectory.

To structure the input data into a format suitable for autoregressive modeling, each trajectory is treated as a discrete sequence of location-based tokens. For each trajectory, we incorporate an ``end of trajectory'' token, denoted as \texttt{<EOT>}, which is appended to each trajectory during training. Given a trajectory sequence $\phi \in \Phi = \{\phi_{1}, \phi_{2}, ..., \phi_{n}\}$ in terms of road links, where $\phi_i$ denotes the $i$-th token in the trajectory. The tokenization process is defined as follows:
\begin{align}
\mathbf{\Phi}_{\text{tokenized}} = \{\phi_1, \phi_2, ..., \phi_n, \texttt{<EOT>}\}
\end{align}

\textcolor{black}{This inclusion of \texttt{<EOT>} serves a dual purpose. First, it acts as a sentinel token to signify the termination of a trajectory. Formally, $\text{if } \phi_i = \texttt{<EOT>} \text{, then trajectory } i+1 \text{ begins}$. This structural sequence tokenization enables the model to distinguish between successive locations, thus avoiding positional leakage across trajectories during batch processing. This prevents the model from interpreting the start of a new trajectory in a batch as a continuation of the previous one, maintaining clear trajectory boundaries during both training and generation. Second, the proposed tokenization supports the generation of diverse and randomized trajectory sequences by clearly marking sequence endpoints, which facilitates autoregressive sampling without conflating distinct OD pairs. As \texttt{<EOT>} is part of the model’s vocabulary, MobilityGPT assigns it a probability at every decoding step. This allows trajectories to terminate at variable positions, enabling the generation of diverse and randomized geospatial sequences while preserving structural coherence with real-world mobility patterns. }

\textcolor{black}{Moreover, this design abstracts away from raw GPS coordinates, allowing the model to learn high-level movement patterns through a discrete vocabulary of meaningful road link identifiers. This abstraction mitigates the impact of noise from GPS sparsity or irregular sampling rates and grounds the representation in topological connectivity rather than continuous space.}

Utilizing this tokenization strategy is integral to our model's ability to capture spatial patterns, differentiate between trajectories, and generate realistic sequences with varied ODs. This formulation contributes to the robustness and versatility of our trajectory generation approach.

\subsection{Map Understanding with Gravity Model and Road Connectivity Matrix}

Transformer-based models capture sequential information within long data sequences. However, human mobility modeling presents distinct challenges, necessitating a comprehensive understanding of geographical attributes and constraints. For instance, the sequence of links does not contain hidden attributes of human mobility, such as regional mobility flows (traveling from eastbound of city to west), traffic rules (must right-turn, no U-turn), and trip length. To address these limitations, we integrate two structured priors from classical mobility modeling—the gravity model and the RCM—into the training process of the transformer.

\textcolor{black}{While both models are well-established in the transportation literature \cite{cabanas2025human, wang2023attention}, our work recontextualizes them as inductive biases within a deep generative model. Specifically, the gravity model is repurposed as a data sampling prior to expose the transformer to representative and balanced regional flows, while the RCM is injected into the final decision layer of the model to constrain predictions to physically feasible road transitions. These strategies collectively guide the transformer to learn geospatial structure without requiring explicit auxiliary data such as maps or network graphs.}

\begin{figure*}[t]
    \centering
    \includegraphics[width=0.9\linewidth]{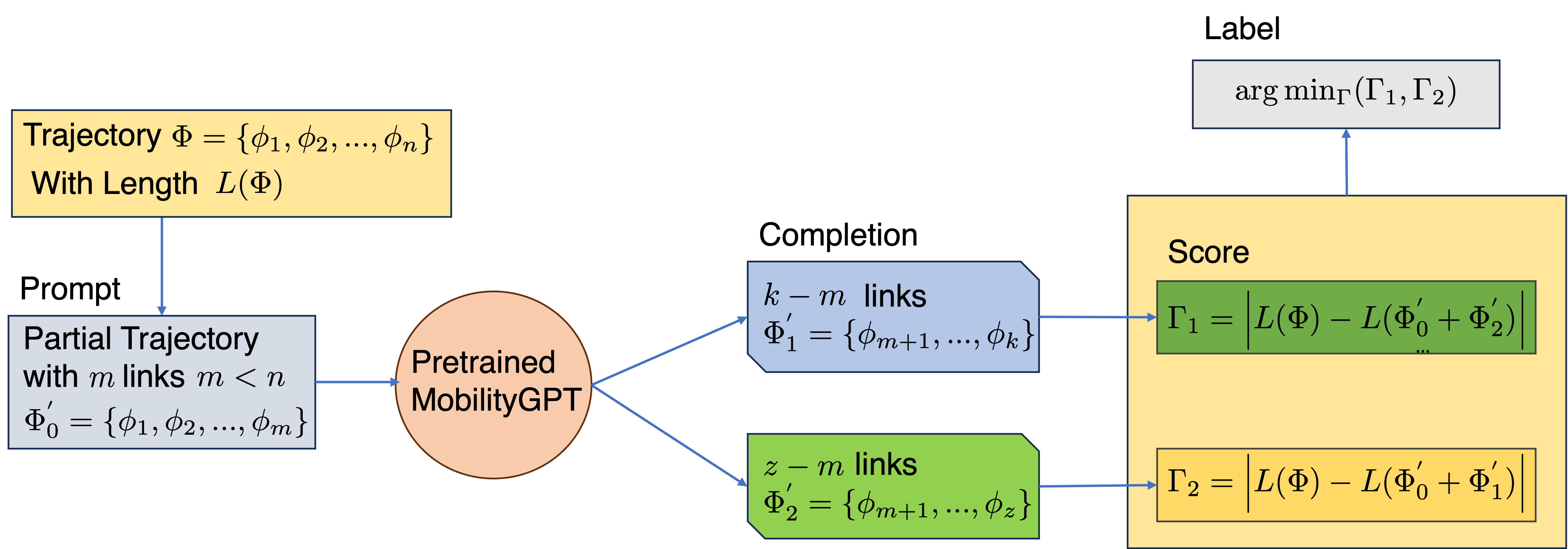}
    % \vskip -0.1in
    \caption{The scheme of fine-tuning dataset generation for Reinforcement Learning from Trajectory Feedback.}
    \label{fig:rltf_dataset}
    % \vskip -0.2in
    \vspace{-0.5cm}
\end{figure*}

\textcolor{black}{{\bf 1) Gravity model.} The area is discretized into regions $r \in \mathcal{R}$, with $\text{RegionW}(r_{x})$ calculated as the number of trajectories starting or ending in the region $r_{x}$. The gravity value $\text{Gravity}(r_{x}, r_{y})$ quantifies traffic flow between origin $r_{x}$ and destination $r_{y}$ regions as:
\begin{equation}
\text{Gravity}(r_{x}, r_{y}) = \frac{\text{RegionW}(r_{x}) \cdot \text{RegionW}(r_{y})}{d^{2}(r_{x}, r_{y})}
\end{equation}
where $d^{2}(r_{x}, r_{y})$ is the squared Euclidean distance between regions. This classical formulation has historically been used to model inter-regional travel demand. In our case, we adopt it as a training-time sampling prior to ensure that the model observes a diverse and flow-consistent distribution of OD pairs. During training, as shown in Figure~\ref{fig:MobilityGPT_overall}, a gravity-aware sampler selects trajectory sequences $\Phi = {\phi_{1}, \phi_{2}, ..., \phi_{n}}$ based on $\text{Gravity}(r_{x}, r_{y})$, where $\phi_{1} \in r_{x}$ and $\phi_{n} \in r_{y}$. This strategy reduces sampling bias toward overrepresented regions and encourages the model to internalize global mobility trends.}

{\bf 2) Road connectivity matrix.} The clarity of road connectivity is often diminished when using GPS trajectories compared to traditional 2D maps. This is because, in some instances, roads that appear connected on a map might not be so in reality due to physical barriers or design differences, such as one-way streets and pedestrian zones. \textcolor{black}{To incorporate real-world structural constraints into sequence generation, we propose the RCM, a sparse binary matrix that reflects the physical adjacency of road links.}

Let $\Phi$ denote the set of training trajectories in terms of road links. The RCM is defined as:
\begin{equation}
RC_{\phi_x,\phi_y} =
\begin{cases}
1 & \text{if } (\phi_x, \phi_y) \text{ is a consecutive pair in } \Phi, \\
 & \quad \text{or} \\
  & \text{if } \phi_x \text{ or } \phi_y \text{ is in the last row/column,} \\
0 & \text{otherwise}.
\end{cases}
\end{equation}
RCM effectively represents the connectivity between different road links. If two links, $\phi_x$ and $\phi_y$, are directly connected in the real world, their corresponding element in RCM is set to 1; otherwise, it is set to 0. \textcolor{black}{This structure reflects only plausible transitions observed in the data. To enforce these structural constraints consistently, we apply the RCM to the transformer's output logits during both training and generation, ensuring that the model only learns and predicts physically plausible road transitions.} Additionally, one row and one column, both filled with 1s, are added to the matrix to facilitate the generation of \texttt{<EOT>}. Specifically, the last linear layer of the transformer, denoted as $F$, is adjusted by the RCM prior to the softmax activation function:
\begin{equation}
F' = F \cdot RC
\end{equation}
\begin{equation}
\text{Softmax}(F') = \frac{e^{F'}}{\sum e^{F'}}
\end{equation}
this integration is visualized in Figure~\ref{fig:MobilityGPT_overall}, showing how the RCM modifies the output of the transformer's last linear layer. By multiplying $F$ with the RCM, we ensure that only the connections between actually connected roads are considered. This process filters out any inaccuracies in road connectivity that might have been introduced by GPS data, thus enabling the model to have a more accurate understanding of the actual road network.

\subsection{Reinforcement learning from trajectory feedback (RLTF)}

Fine-tuning is essential for sequence generative models \citep{zhang2023instruction}. Pre-trained transformer architectures provide a foundation, but fine-tuning customizes models to specific objectives, such as similarity of trajectories, enabling them to learn complex patterns and nuances. This process is particularly vital in applications like human mobility, where capturing intricate movement patterns is crucial. 
Building on this, we propose a novel approach to construct a trajectory-aware preference dataset that combines established fine-tuning strategies with a new preference feedback dataset - all without requiring human labeling. This enables the model to self-improve based on its output feedback, minimizing human intervention and addressing scalability and bias challenges \citep{ouyang2022training}.

Reinforcement learning has proven to be a powerful approach for capturing the long-term dependencies and spatiotemporal characteristics inherent in human mobility modeling, owing to the complexity of this task  \citep{reed2022generalist}. The proposed approach, incorporating our novel preference dataset with reinforcement learning, termed Reinforcement Learning from Trajectory Feedback (RLTF), in an automated feedback process, involves several key steps. First, a reward dataset is created by pairing prompts (partial trajectories) with compilations (complete trajectories generated by the model) and assigning scores based on the similarity of the compilation lengths to the original trajectory lengths (see Figure \ref{fig:rltf_dataset}). This reward dataset is then utilized to train a reward model $\mathbf{U}$ through supervised learning. Subsequently, the trained reward model $\mathbf{U}$ is employed to fine-tune the MobilityGPT model through a Proximal Policy Optimization (PPO) policy $\pi$, effectively leveraging reinforcement learning without requiring human feedback.   

\textcolor{black}{The reward model $\mathbf{U}$ is a crucial component of our RLTF framework. We adopted the same lightweight and efficient model architecture as MobilityGPT itself for the reward model. This design choice ensures computational efficiency and a seamless integration with the main model, allowing the reward signal to be accurately modeled without introducing significant overhead. By using a shared architecture, we can effectively capture the complexities of the reward function while keeping the overall system resource-friendly.}

{\bf Preference dataset construction for RLTF.} Instead of relying on human feedback in fine-tuning of MobilityGPT, we propose using quantitative metrics to construct the preference dataset. Given a partial trajectory prompt, we generate two complete trajectories and assess their similarity to the actual trajectory by assigning a trajectory similarity score based on the relative lengths of the generated and reference trajectories from the training set \textcolor{black}{(See Step 2 in Figure~\ref{fig:MobilityGPT_overall})}. This score measures how well the generated trajectories match the characteristics, particularly length, of the training data, serving as a proxy for human preference without requiring human labeling.

\begin{figure}[t]
    \centering
    \includegraphics[width=0.9\linewidth]{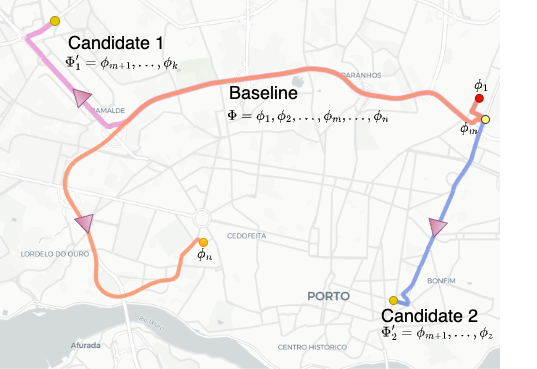}
    \caption{Example of a preference dataset, showing a sample input alongside two completed trajectory pairs. Candidate 1’s trajectory demonstrates greater similarity to the Baseline trajectory compared to Candidate 2.}
    \label{fig:rltf_example}
\end{figure}

The process, as described in Figure~\ref{fig:rltf_dataset}, begins with a full trajectory $\Phi = {\phi_1, \phi_2, \ldots, \phi_n}$ of length $L(\Phi)$. The first $m$ road-links $\Phi'_0 = {\phi_1, \phi_2, \ldots, \phi_m}$ serve as a prompt for MobilityGPT to predict two possible continuations: $\Phi'_1 = {\phi_{m+1}, \ldots, \phi_k}$ and $\Phi'_2 = {\phi_{m+1}, \ldots, \phi_z}$. These are concatenated with $\Phi'_0$ to form two complete trajectories, evaluated using $\Gamma_1 = |L(\Phi) - L(\Phi'_0 + \Phi'_2)|$ and $\Gamma_2 = |L(\Phi) - L(\Phi'0 + \Phi'1)|$. The trajectory with lower $\Gamma$ is labeled $\Phi_{\text{chosen}}$, and the other $\Phi_{\text{rejected}}$, forming a preference dataset for RLTF. By leveraging this proximity-based metric, we eliminate the need for human labeling, streamlining the dataset creation process and potentially mitigating biases introduced by human subjectivity. MobilityGPT can be generalized to different fine-tuning strategies with the proposed preference dataset. \textcolor{black}{In Figure \ref{fig:rltf_example}, we exemplified one step of the preference dataset creation schema with an actual sample trajectory using the Porto dataset \cite{jiang2023continuous}. }

{\bf Reward model training.} 
A Reward (preference) Model $\mathbf{U}$ is trained using the proposed preference dataset to predict the probability that a given trajectory is chosen over a rejected alternative \textcolor{black}{(See Step 2 in Figure~\ref{fig:MobilityGPT_overall})}. The loss function is formulated as follows:
\begin{equation}
-\log(\sigma(\mathbf{U}(\Phi_{\text{chosen}}) - \mathbf{U}(\Phi_{\text{rejected}})))
\end{equation}
The goal of the loss function is to maximize the probability that the chosen trajectory has a higher score, thereby encouraging the model to assign a larger difference between $\Phi_{\text{chosen}}$ and $\Phi_{\text{rejected}}$. As the training progresses, the model becomes more capable of recognizing the characteristics of trajectories that align with the preferred trip length. 

{\bf Fine-tuning with reinforcement learning.} \textcolor{black}{Given the pre-trained MobilityGPT model (Step 1) and supervised trained reward model (Step 2), we train an RL policy that optimizes the MobilityGPT model for generating higher-quality synthetic trajectories ss depicted in Step 3 of Figure~\ref{fig:MobilityGPT_overall}.} Since policy learning relies on the reward that policy receives from the environment, the reward model provides feedback for RL policy optimization. We followed a general policy optimization method as in \citep{ziegler2019fine}. The RL agent learns a fine-tuning policy $\pi$, which is optimized with the well-known Proximal Policy Optimization (PPO) method \citep{schulman2017proximal}. The reward of policy $\pi$ consists of two terms: reward model logits given the prompt $\Phi'_p$ and compilation $\Phi'_c$ pair received from the MobilityGPT as $\mathbf{U}(\Phi'_p, \Phi'_c)$ and policy shift constraint in terms of the (Kullback–Leibler) KL divergence given the baseline policy $\pi$. The full reward can be represented as follows:
\begin{equation}
\text{R}(\Phi'_p, \Phi'_c) = \mathbf{U}(\Phi'_p, \Phi'_c) - \beta \log\left(\frac{\pi^{\text{PPO}}_{\theta'}(\Phi'_c|\Phi'_p)}{\pi^{\text{base}}(\Phi'_c|\Phi'_p)}\right).
\end{equation}

It is worth emphasizing that the neural network architecture of the reward model $\mathbf{U}$ can be simplified to enhance efficiency and interpretability. Moreover, there is the option to incorporate an online fine-tuning stage, where the policy $\pi$ and the reward model $\mathbf{U}$ are periodically re-trained. However, in our experimental assessments, we did not delve into the performance of the retraining process. This aspect and the formal exploration of privacy-preserving ML models are left for future research endeavors.

\section{Experiments}
\label{s:results}

MobilityGPT's effectiveness is validated through extensive experiments on real-world datasets against recent benchmarks. We performed all the experiments on a single NVIDIA TITAN RTX GPU with 24GB memory. The source code of MobilityGPT is publicly available. The preprint version of this paper is available on arXiv at \citep{haydari2024mobilitygpt}.
\footnote{\url{https://github.com/ammarhydr/MobilityGPT}}

\subsection{Experimental setup}

We have used two datasets of GPS trajectories from taxi drivers in Porto, Portugal, and Beijing, China. The Porto dataset has a sampling period of $15$ seconds, whereas the Beijing dataset has a sampling period of $1$ minute. \textcolor{black}{Both datasets were preprocessed and map-matched following the protocol described by \cite{jiang2023continuous}. Specifically, the authors filtered out abnormal trajectories by removing those shorter than five road links or containing loops. After preprocessing, the Porto dataset comprises $695,085$ unique trips across $11,095$ road links, while the Beijing dataset includes $956,070$ trajectories spanning $40,306$ road links.}

\textcolor{black}{We chose these two datasets for two primary reasons. First, they are widely recognized and frequently used as benchmark datasets in the field of human mobility modeling, particularly in recent works such as TS-TrajGen \citep{jiang2023continuous} and DiffTraj \citep{zhu2024difftraj}. This established use allows for a direct and fair comparison of our model's performance against state-of-the-art benchmarks. Second, the datasets represent distinct geographic regions and mobility patterns. By validating our model on data from different continents with varying urban structures, we aim to demonstrate its robustness and ability to accurately model diverse human mobility characteristics, irrespective of geographical location or a specific urban context.}

\subsection{Benchmark models}
\label{app:benchmarks}
We performed experiments on original implementation hyperparameters for all benchmark models using their publicly available source codes. We experimented with similar structured generative models with our proposed MobilityGPT model, including discrete and continuous structure formats.
\begin{itemize}
    \item \textcolor{black}{\textbf{Random Walk}: This baseline randomly selects a starting point on the road network and performs a random walk guided by visitation counts of the road segments (locations). It operates solely on the road graph structure without learning from the data distribution.}
    
    \item \textcolor{black}{\textbf{Mobility Markov Chain (MMC)} \cite{gambs2012next}: This model treats road segments as discrete states and constructs a transition probability matrix to capture first-order dependencies between segments. The next segment is sampled based on the learned transition probabilities.}
    
    \item \textbf{SeqGAN} \citep{yu2017seqgan}: SeqGAN is a state-of-the-art sequential generative model that trains trajectories in discrete sequences with RL policy gradient methods using Monte Carlo sampling.
    \item \textbf{LSTM-TrajGAN} \citep{rao2020lstm}: This is a tabular LSTM model by learning the normalized GPS location divergences with respect to the center of the dataset.
    \item \textbf{TS-TrajGen} \citep{jiang2023continuous}: TS-TrajGen is a method based on GANs that incorporates a modified $A*$ path search algorithm. This adjustment ensures spatial continuity in the generated data. Additionally, the method includes topological constraints to align the generated trajectories with existing road networks, ensuring road matching. TS-TrajGen requires fixed OD pairs as inputs. We use OD pairs from the real training set as inputs during inference.
    \item \textbf{DiffTraj} \citep{zhu2024difftraj}:  is the most recent and promising generative model for mobility trajectories. DiffTraj employs a diffusion model in the U-Net structure for generating GPS trajectories.
\end{itemize}

\subsection{Evaluation metrics}
\label{app:metrics}

\textcolor{black}{To evaluate the performance of the generative models, we require robust comparison metrics that capture a broad spectrum of mobility characteristics and effectively quantify the quality of the generated trajectories relative to real ones. The selected metrics assess whether the synthetic trajectories preserve both local transition patterns—i.e., movements between adjacent road segments—and global route structures that reflect realistic and rational human mobility behavior in urban environments. In this study, we adopted four widely used evaluation metrics, along with our proposed gravity and connectivity utility metrics, as introduced in prior works \citep{hasan2013spatiotemporal, gursoy2018utility}. These metrics have become standard tools for assessing the utility of synthetic mobility data generation methods \citep{haydari2022differentially, zhu2024difftraj}. Given that MobilityGPT introduces randomness and can alter trajectory patterns, direct one-to-one trajectory comparisons are not applicable in such generative settings \citep{kapp2023generative}.}

\textbf{Query Error}: The query error, mainly used for evaluating data synthesis algorithms, is a popular metric for synthetic data generation models. We use spatial counting queries in the form of ``the number of trajectories passing through a certain road-link".  Given road network $D(V,E)$, we uniformly sampled $500$ road links from the road network. Then, the normalized absolute difference between the number of real and synthetic trajectories passing through each of these links is computed by the following:
\begin{equation}
 \hbox {QE}(f(\Sigma)) = \frac{|f(\Psi)- f(\Sigma)|}{\max \left\{ f(\Psi), s\right\} }, 
\end{equation}
where $f(\Psi)$ and $f(\Sigma)$ stand for the query outcome from the original and synthetic trajectories, respectively, and $s$ is sanity bound for mitigating the effect of the extremely small selective queries. We specified the sanity bound $s$ as $1\%$ of the users. 

Apart from the query error, the main utility metrics inspect the distributional similarities for synthetic trajectories with respect to real trajectories that train the generative model. Jensen-Shannon divergence (JSD) is a well-known similarity metric mainly used for measuring the similarity of two probability distributions \citep{lin1991divergence}. We employ JSD for several human mobility characteristics. 

We can look at overall distributions of the generated trajectories $\Lambda_{gen}$ and real trajectories $\Lambda_{real}$, and JSD can provide a summary statistic for a pair of distributions. These are the main distributions we extract from the $\Lambda_{gen}$ and $\Lambda_{real}$:
\begin{itemize}
    \item \textbf{OD}: This metric evaluates the origin-destination similarity between two datasets for the overall characteristics preserved in terms of OD links.
    \item \textbf{Trip Length}: This metric considers the travel distances of trajectories by using the total length of link trajectories. 
    \item \textbf{Radius}: In human mobility patterns, the spatial range of daily movements is an important metric. We examined the user's radius of gyration within the controlled road network area. 
    \item \textbf{Gravity}: The gravity summary metric examines the impact of the gravity sampling model that was introduced for MobilityGPT at high-level mobility patterns with the same JSD method. 
\end{itemize}

In addition to divergence metrics, we introduce a new metric that measures link trajectory connectivity as a percentage. The connectivity metric quantifies the percentage of trajectories that are fully connected from origin to destination, owing to the RCM that samples the next road link conditioned on the connected road links, thereby capturing the underlying road network topology and connectivity patterns. We employed the connectivity metric in our ablation studies to assess the impact of various components on link trajectory connectivity.

\subsection{Model Details}
\label{s:params}

MobilityGPT utilizes a minimal version of the standard decoder-only transformer architecture \citep{radford2019language}, built upon the minimal GPT implementation from \citep{karpathy_mingpt}. Table \ref{tab:hyperparams} summarizes the important model and training parameters employed in this work. Since the Porto and Beijing datasets exhibit different trajectory lengths and characteristics, we utilize distinct block sizes for training and maximum trajectory lengths for synthetic trajectory generation. However, we employ the same set of hyperparameters across both datasets for all training stages, including the reward model and PPO training.

\begin{table}[t]
\centering
\caption{Model and training parameters utilized in MobilityGPT}
\label{tab:hyperparams}
    \ra{1.0}
    \begin{tabular}{cc}
        \Xhline{1pt} % Thick line at the top
        \hline
        \hline
        Parameter & Value\\
        \hline
        N. layers & 6\\
        N. heads & 4 \\ 
        N embeddings & 64 \\ 
        Batch size & 64 \\ 
        Learning rate & 1e-5 \\
        Training size & 80\% \\
        Test size & 20\% \\
        Training steps & 3000 \\        
        Block size (Porto) & 300\\
        Block size (Beijing) & 60\\
        Max. Traj. length (Porto) & 278 \\
        Max. Traj. length (Beijing) & 60 \\
        \hline
        \Xhline{1pt}
    \end{tabular}
\end{table}

\subsection{Comparison with benchmark generative methods}

\begin{table*}[t]
\centering
\fontsize{9.5pt}{10pt}\selectfont
%\vskip -0.1in
\ra{1.3}
\caption{Comparison of the utility metrics with benchmark studies. Bold and underlined results show the best and second-best results, respectively. For all metrics, lower values indicate better performance. While query error relies on normalized error rates for most visited places, other metrics use the Jensen-Shannon distributional divergence statistic.}
\label{tab:models_comparison} 
\vskip 0.1in
\begin{tabular}{@{}lccccc|ccccc@{}}
\Xhline{1pt} % Thick line at the top
\hline
\hline
 & \multicolumn{5}{c}{Porto-Taxi} & \multicolumn{5}{|c}{BJ-Taxi}  \\
\cmidrule{2-6}  \cmidrule{7-11}
Methods & Query E. & OD & Trip L. & Radius & Gravity & Query E. & OD & Trip L. & Radius & Gravity \\ 

\cmidrule{1-1} \cmidrule{2-6}  \cmidrule{7-11}
Random Walk &1.729&0.375&0.739&0.393&0.307&0.105&0.167&0.473&0.187&0.133\\
MMC &1.409&0.210&0.742&0.393&0.307&0.212&0.167&0.550&0.465&0.291\\
SeqGAN  & \textbf{0.138} & \underline{0.116} & 0.220 & 0.191 & 0.326 &\textbf{0.063} &  0.296 & \underline{0.162} & 0.275 & 0.120 \\
LSTM-TrajGAN  & 0.234 & 0.254 & 0.435 & 0.425 & 0.277 &  0.078 & 0.182 & 0.398 & 0.365 &0.219 \\
TS-TrajGen  & 0.309 & 0.120 & 0.161 & \underline{0.115} & \underline{0.248} & 0.277 &  \underline{0.105} & 0.570 & 0.485  &0.336 \\
DiffTraj  & 0.267 & 0.177 & \underline{0.126} & 0.123 & 0.252 & 0.074 & 0.146 & 0.158 & \underline{0.123} & \underline{0.114} \\
\textbf{MobilityGPT} & \underline{0.144} & \textbf{0.114} & \textbf{0.124} & \textbf{0.107} & \textbf{0.225} & \underline{0.066} & \textbf{0.099} & \textbf{0.123} & \textbf{0.102}&\textbf{0.105} \\ 
% \midrule
\hline
\hline
\Xhline{1pt}

\end{tabular}
\vskip -0.1in
\end{table*}

Table~\ref{tab:models_comparison} summarizes the performance of MobilityGPT and benchmark models. We sampled $5000$ random trajectories from the test set and generated same number of synthetic trajectories. The results show the significant advantage of MobilityGPT over existing state-of-the-art models for trajectory generation. Across both Porto-Taxi and BJ-Taxi, MobilityGPT consistently outperforms all baseline models, achieving the best or second-best performance on multiple utility metrics.

\textcolor{black}{
The results in Table~\ref{tab:models_comparison} demonstrate that MobilityGPT consistently outperforms baseline models across both the Porto and Beijing datasets and all utility metrics. Rule-based methods such as Random Walk and MMC show relatively high divergence scores, indicating poor alignment with real-world distributions. While deep generative models like SeqGAN and DiffTraj achieve competitive results on some metrics, they lack consistency across all aspects of trajectory modeling. In contrast, MobilityGPT achieves the lowest JS divergence across all categories on both datasets—for instance, reducing OD distribution divergence to $0.114$ (Porto) and $0.099$ (Beijing), outperforming the next best scores of $0.116$ and $0.105$, respectively. This performance reflects the combined impact of MobilityGPT’s autoregressive transformer architecture and its gravity-based sampling strategy, which better captures global mobility patterns. By grounding generation in structural road connectivity and learned spatio-temporal dynamics, MobilityGPT shows strong potential for applications in transportation, urban planning, and mobility analytics.
}

\begin{figure*}[ht!]
    \centering
    \subfigure[Distribution of number of points per trajectory for the Porto-Taxi dataset.]{\includegraphics[width=0.45\textwidth]{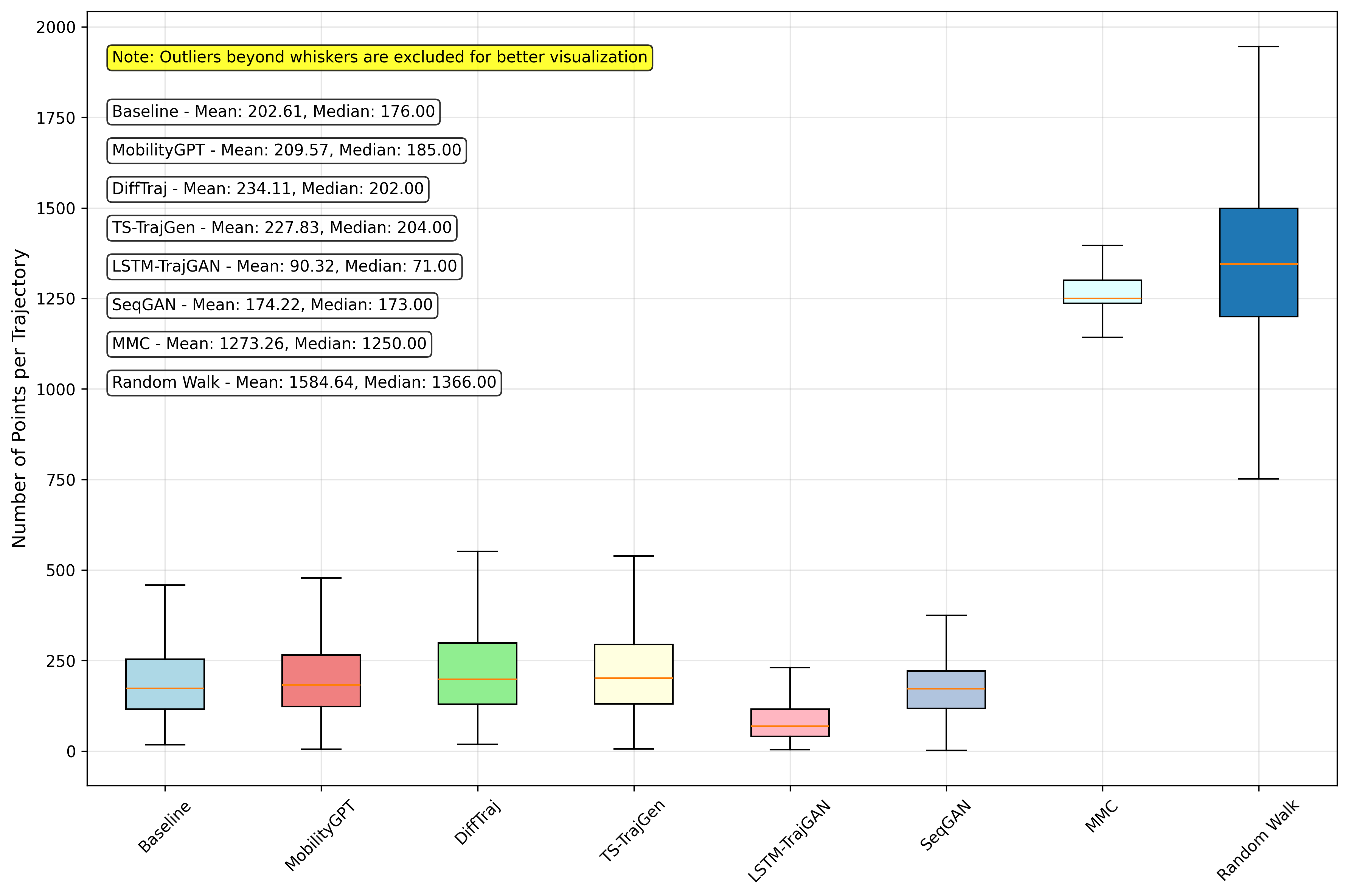} \label{f:porto_points}}
    \hfill
    \subfigure[Distribution of number of points per trajectory for the BJ-Taxi dataset.]{\includegraphics[width=0.45\textwidth]{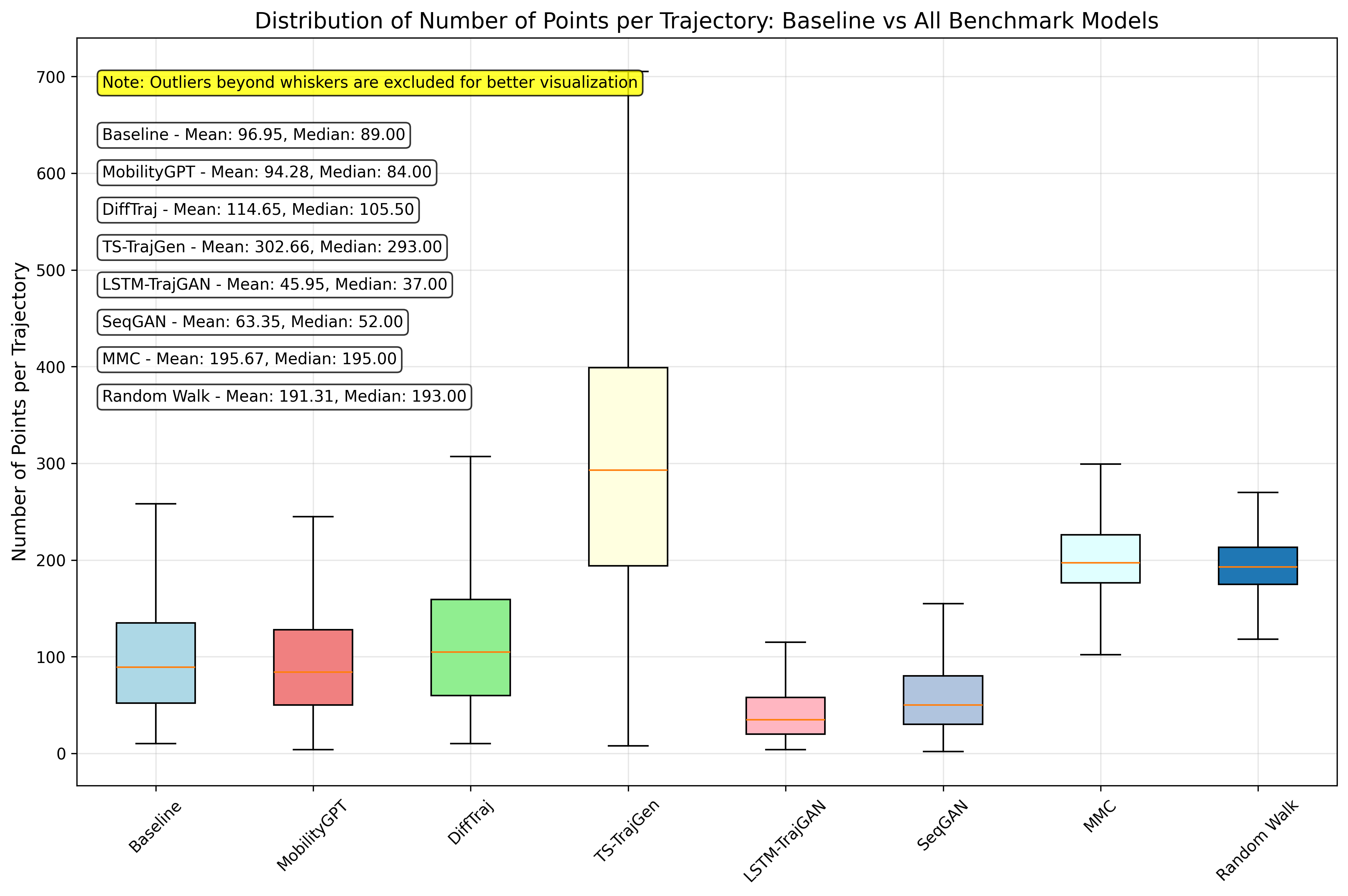} \label{f:bj_points}}
    \vspace{1cm}
    \subfigure[Distribution of inter-point distance for the Porto-Taxi dataset.]{\includegraphics[width=0.45\textwidth]{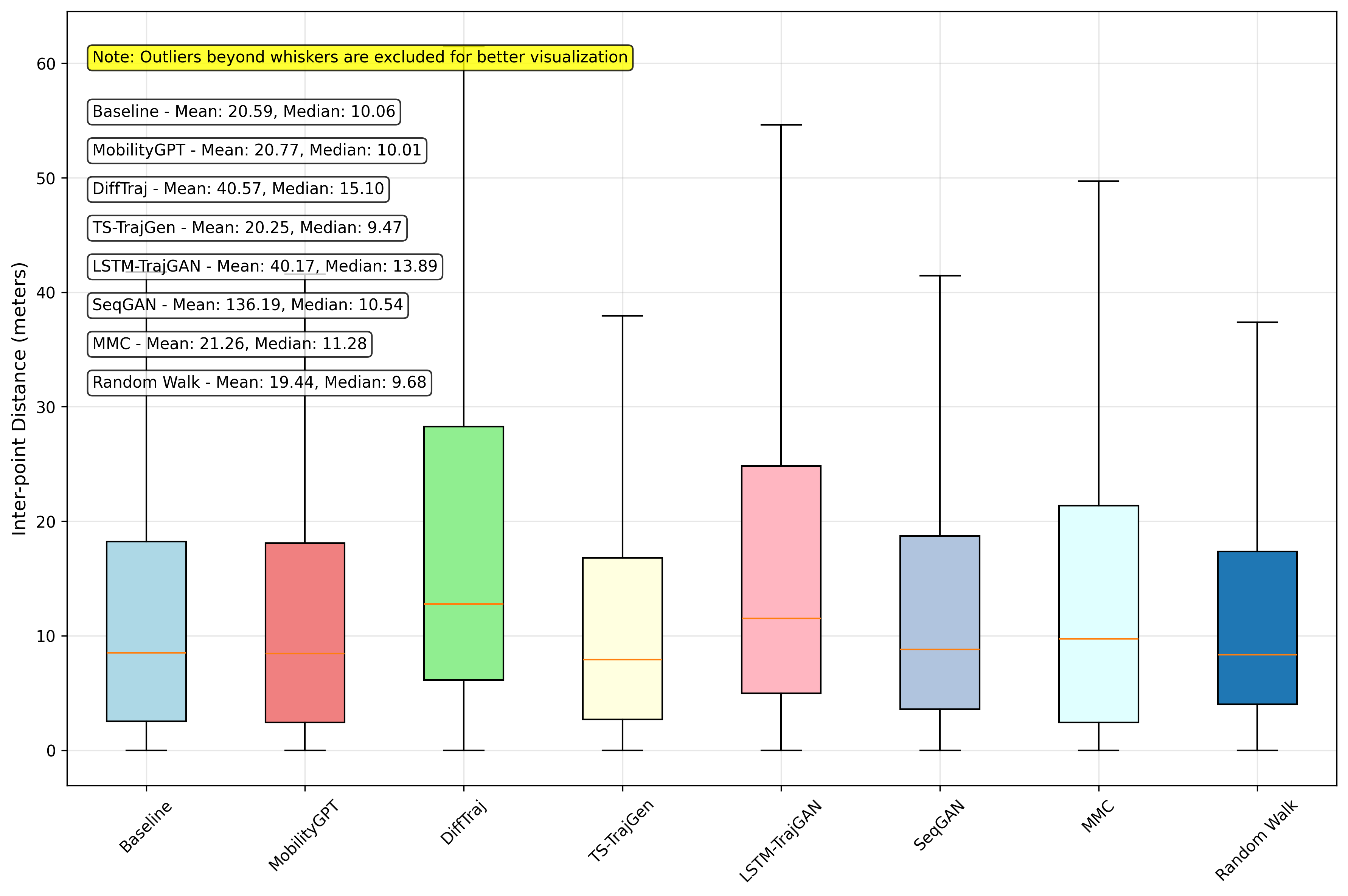} \label{f:porto_distance}}
    \hfill
    \subfigure[Distribution of inter-point distance for the BJ-Taxi dataset.]{\includegraphics[width=0.45\textwidth]{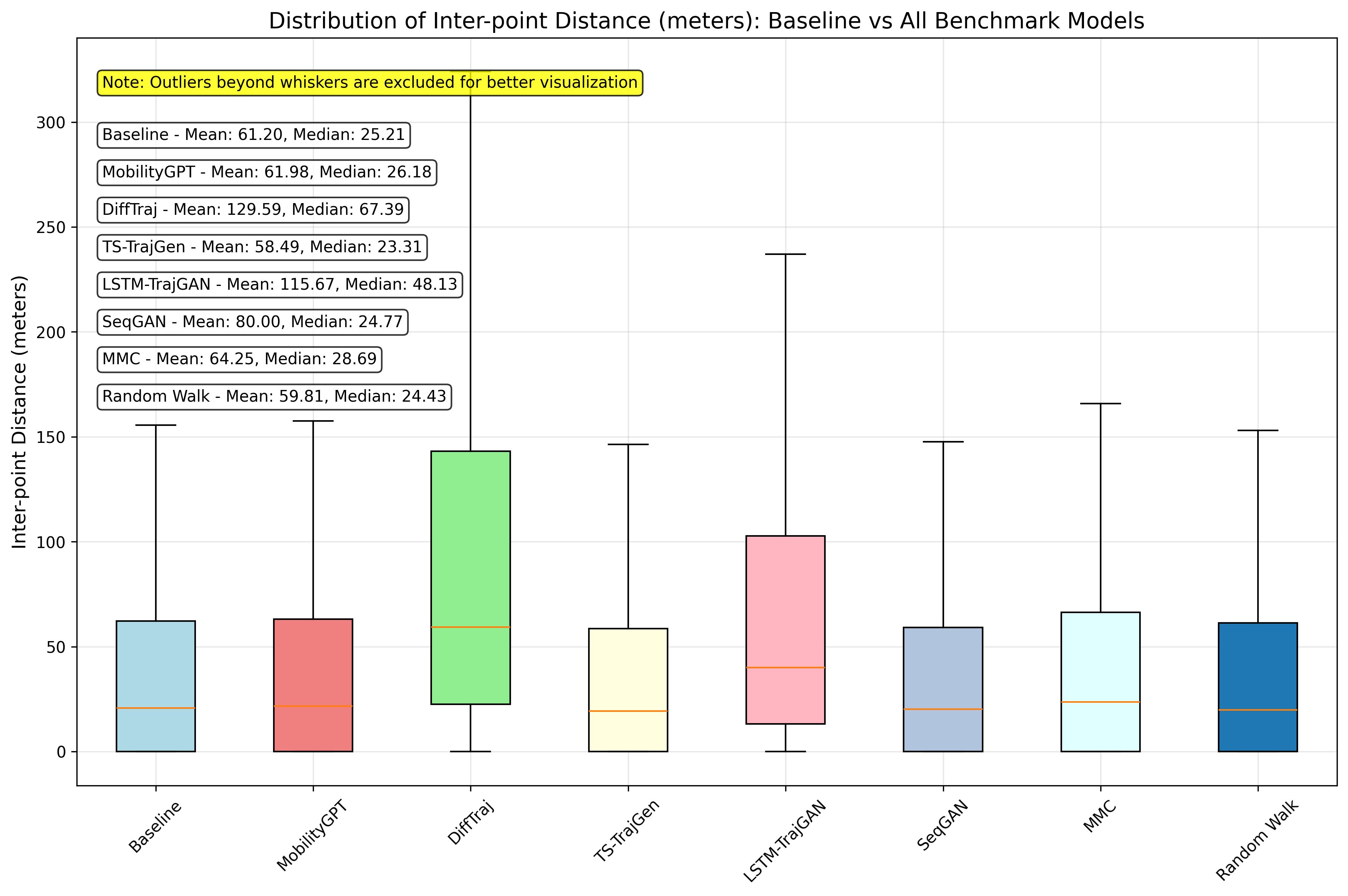} \label{f:bj_distance}}
    \caption{Comparison of different trajectory generation models on the Beijing and Porto datasets. The plots show the distribution of the number of points per trajectory and the inter-point distance.}
    \label{f:all_box_plots}
\end{figure*}

\textcolor{black}{The box plots in Figure~\ref{f:all_box_plots} illustrate the performance of MobilityGPT and several other models across two key metrics: the number of points per trajectory and inter-point distance. Analysis of the Number of Points per Trajectory reveals that MobilityGPT's performance is consistently strong. On the Porto dataset (Figure~\ref{f:porto_points}), its median of $185.00$ is very close to the Baseline's $176.00$, demonstrating that MobilityGPT effectively replicates the trajectory length characteristics of this dataset. Similarly, for the Beijing dataset (Figure~\ref{f:bj_points}), MobilityGPT's median of $84.00$ is remarkably close to the Baseline's $89.00$. This indicates that MobilityGPT successfully replicates the fundamental characteristic of trajectory length from the original datasets, a crucial aspect of generating realistic, synthetic mobility data. In contrast, models like TS-TrajGen, MMC, and Random Walk show significantly different distributions, suggesting they generate trajectories that are either excessively long or follow a different statistical pattern than the baseline.}

\textcolor{black}{Regarding Inter-point Distance, MobilityGPT also demonstrates a high degree of fidelity to the baseline data. For the Porto dataset (Figure~\ref{f:porto_distance}), the results are highly aligned, with MobilityGPT's mean of $20.77$ meters and median of $10.01$ meters nearly identical to the Baseline's $20.59$ meters and $10.06$ meters, respectively. On the Beijing dataset (Figure~\ref{f:bj_distance}), MobilityGPT's mean of $61.98$ meters and median of $26.18$ meters are highly comparable to the Baseline's mean of $61.20$ meters and median of $25.21$ meters. These results underscore that MobilityGPT not only generates trajectories of the correct length but also accurately captures the realistic step distances between consecutive points. This is a significant finding, as models like DiffTraj and SeqGAN exhibit a much wider variance or different median values, suggesting they may struggle to accurately represent the typical movement patterns found in real-world data. Overall, the consistent alignment of MobilityGPT's metrics with the Baseline across both datasets confirms its effectiveness in generating high-fidelity, synthetic trajectory data.}

\begin{figure*}[tb]
\centering
\vskip -0.1in
\subfigure[Original Trajectories]{\includegraphics[width=0.32\textwidth]{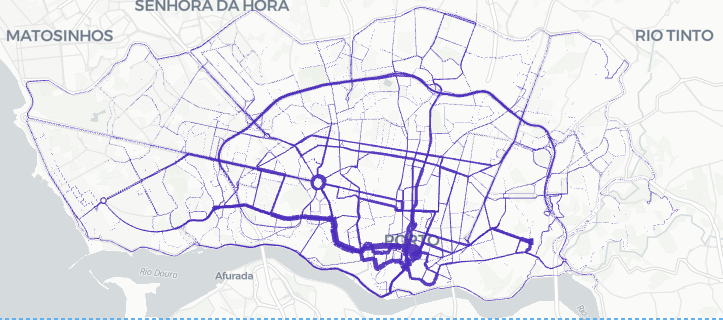} \label{f:org}}
\subfigure[MobilityGPT]{\includegraphics[width=0.32\textwidth]{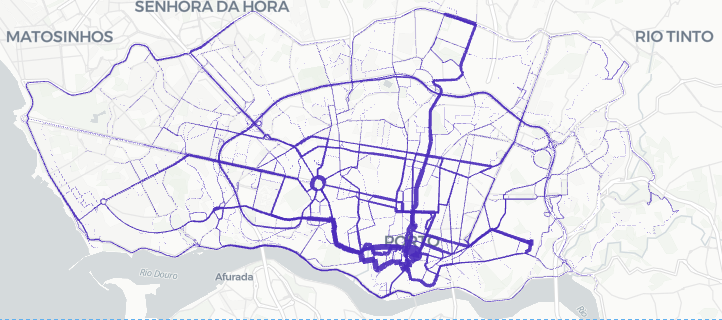} \label{f:MobilityGPT}}
\subfigure[TS-TrajGen]{\includegraphics[width=0.32\textwidth]{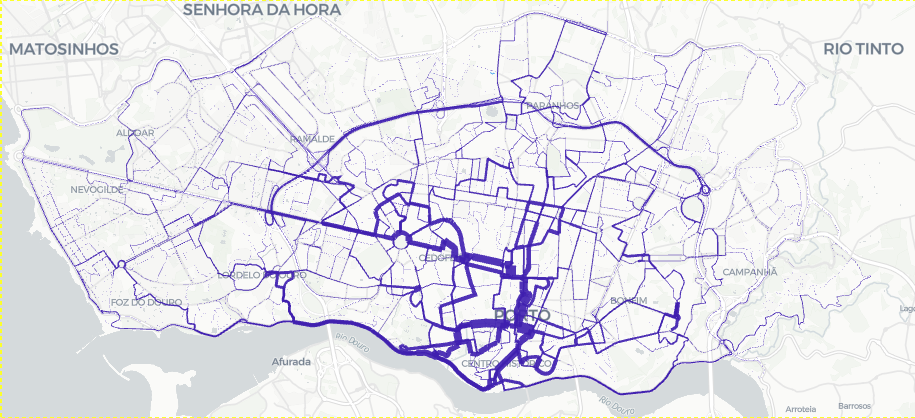} \label{f:ts-trajgen}} \hfill
\subfigure[SeqGAN] {\includegraphics[width=0.32\textwidth]{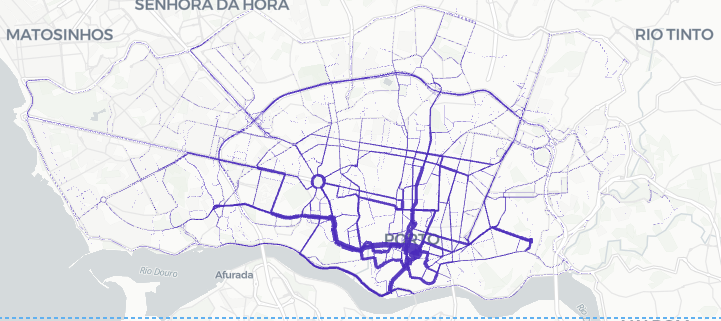} \label{f:seqgan}}
\subfigure[LSTM-TrajGAN] {\includegraphics[width=0.32\textwidth]{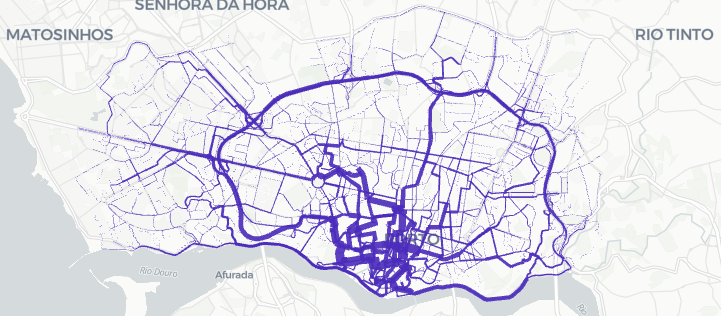} \label{f:lstm-traj}}
\subfigure[DiffTraj] {\includegraphics[width=0.32\textwidth]{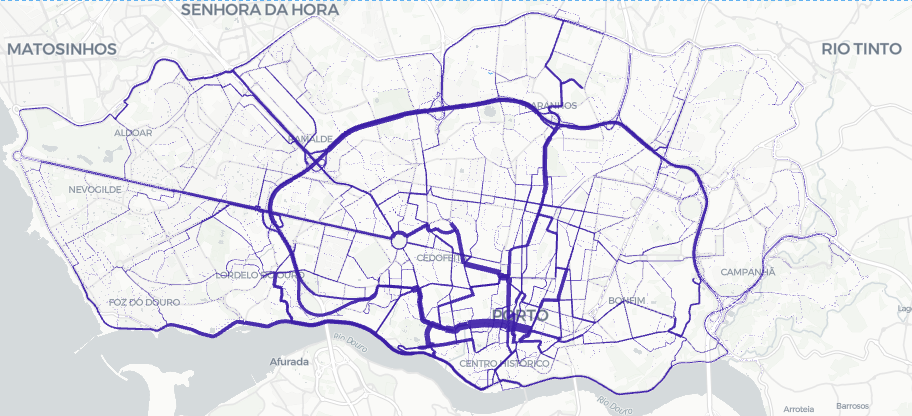} \label{f:difftraj}}
\vskip -0.1in
\caption{Visual representation of the geospatial density of different methods. MobilityGPT exhibits the highest similarity to the original trajectories.}
\label{f:VisualComp}
% \vskip -0.2in
\vspace{-0.5cm}
\end{figure*}

\subsection{Ablation study}

\textcolor{black}{As MobilityGPT incorporates three key enhancements to the standard GPT model for capturing human mobility patterns — gravity model, RCM, and RLTF — we conducted ablation experiments to evaluate their individual and combined contributions. Results, presented in Table~\ref{tab:ablation_study_Porto}a and \ref{tab:ablation_study_BJ}b, show that removing the gravity model and RCM leads to noticeable performance declines. The gravity model plays a critical role in modeling trip lengths, offering substantial improvements even when used in isolation. Without the RCM, MobilityGPT struggles to maintain spatial connectivity in the generated trajectories. By conditioning the next-link prediction on actual road connections, RCM effectively enforces geospatial constraints. Furthermore, fine-tuning MobilityGPT with RLTF improves synthetic trajectory quality—particularly trip length—since the preference dataset is optimized for that objective. These results highlight how each component contributes uniquely to different aspects of mobility modeling, and together, they deliver comprehensive improvements across all evaluation metrics.}

\begin{table}[t]
\centering
\caption{Ablation study on Porto (a) and Beijing (b) datasets examining the impact of different components: RCM, gravity model, and RLTF. Lower values are better for OD and Trip Length, while higher values are better for Connectivity.}
\begin{minipage}{0.5\textwidth}
    \centering
    \ra{1.0}
    \label{tab:ablation_study_Porto}
    \begin{tabular}{cC{13mm}cccc}
        \Xhline{1pt} % Thick line at the top
        \hline
        \hline
        RCM & Gravity Model & RLTF & OD & Trip L. & Connectivity\\
        \hline
        \textcolor{red}{\xmark} & \textcolor{red}{\xmark} & \textcolor{red}{\xmark} & 0.123 & 0.154 & 0.87 \\
        \textcolor{red}{\xmark} & \textcolor{darkgreen}{\cmark} & \textcolor{red}{\xmark} & 0.122 & 0.146 & 0.83 \\
        \textcolor{darkgreen}{\cmark} & \textcolor{red}{\xmark} & \textcolor{red}{\xmark} & 0.120 & 0.147 & 1.0 \\
        \textcolor{darkgreen}{\cmark} & \textcolor{darkgreen}{\cmark} & \textcolor{red}{\xmark} & 0.114 & 0.140 & 1.0 \\
        \textcolor{darkgreen}{\cmark} & \textcolor{darkgreen}{\cmark} & \textcolor{darkgreen}{\cmark} & \textbf{0.114} & \textbf{0.124} & \textbf{1.0}\\
        \hline
        \hline
        \Xhline{1pt}
    \end{tabular}
    \vspace{0.2in}
    \text{(a)}
\end{minipage}%
\hfill
\begin{minipage}{0.5\textwidth}
    \centering
    \ra{1.0}
    \label{tab:ablation_study_BJ}
    \begin{tabular}{cC{13mm}ccC{13mm}c}
        \Xhline{1pt} % Thick line at the top
        \hline
        \hline
        RCM & Gravity Model & RLTF & OD & Trip L. & Connectivity\\
        \hline
        \textcolor{red}{\xmark} & \textcolor{red}{\xmark} & \textcolor{red}{\xmark} & 0.102 & 0.201 & 0.330  \\
        \textcolor{red}{\xmark} & \textcolor{darkgreen}{\cmark} & \textcolor{red}{\xmark} & 0.103& 0.151& 0.367 \\
        \textcolor{darkgreen}{\cmark} & \textcolor{red}{\xmark} & \textcolor{red}{\xmark} & \textbf{0.093}& 0.145& 1.0 \\
        \textcolor{darkgreen}{\cmark} & \textcolor{darkgreen}{\cmark} & \textcolor{red}{\xmark} & 0.099 &0.131& 1.0 \\
        \textcolor{darkgreen}{\cmark} & \textcolor{darkgreen}{\cmark} & \textcolor{darkgreen}{\cmark} & 0.099 & \textbf{0.123} & \textbf{1.0}\\
        \hline
        \hline
        \Xhline{1pt}
    \end{tabular}
    \vspace{0.1in}
    \text{(b)}
\end{minipage}
\end{table}

\textcolor{black}{We compare our proposed fine-tuning approach, RLTF, with two alternative methods: Supervised Fine-Tuning (SFT) and Direct Preference Optimization (DPO)\citep{rafailov2024direct}. Their performance is summarized in Table\ref{tab:ablation_study_RL}. The results show that RLTF is more effective in capturing the complexities of human mobility and outperforms the other fine-tuning strategies. While all methods perform similarly in terms of OD prediction, RLTF achieves significantly lower error in modeling trip length. This improvement suggests that RLTF better captures the nuanced relationship between OD pairs and their associated trip lengths. Its superior performance can be attributed to its use of reward signals that explicitly incorporate spatial distribution and trip length objectives. In contrast, SFT and DPO primarily optimize for input-output matching, which may overlook these deeper patterns in mobility behavior.}

\begin{table}[h]
\centering
\caption{Ablation study on different fine-tuning strategies for MobilityGPT. The metrics use the Jensen-Shannon distributional divergence statistic.}
    \ra{1.0}
    \label{tab:ablation_study_RL}
    \begin{tabular}{ccc>{\columncolor{lightgray}}c>{\columncolor{lightgray}}C{13mm}}
        \Xhline{1pt} % Thick line at the top
        \hline
        \hline
        SFT & DPO & RLTF & OD & Trip Length \\
        \hline
        \textcolor{darkgreen}{\cmark} &  &  & 0.116 & 0.171 \\
        & \textcolor{darkgreen}{\cmark} &  & 0.115 & 0.163 \\
        & & \textcolor{darkgreen}{\cmark} & \textbf{0.114} & \textbf{0.124} \\
        \hline
        \hline
        \Xhline{1pt}
    \end{tabular}
\end{table}

\subsection{Effect of temperature on MobilityGPT inferencing}

The temperature parameter in LLMs, including MobilityGPT, influences the randomness in generating outputs. At lower temperatures, the model behavior is more deterministic, often selecting the most likely next token. As the temperature increases, the model introduces greater randomness, leading to a diverse array of outputs.

\begin{figure}[h]
    \centering
    \includegraphics[width=0.9\linewidth]{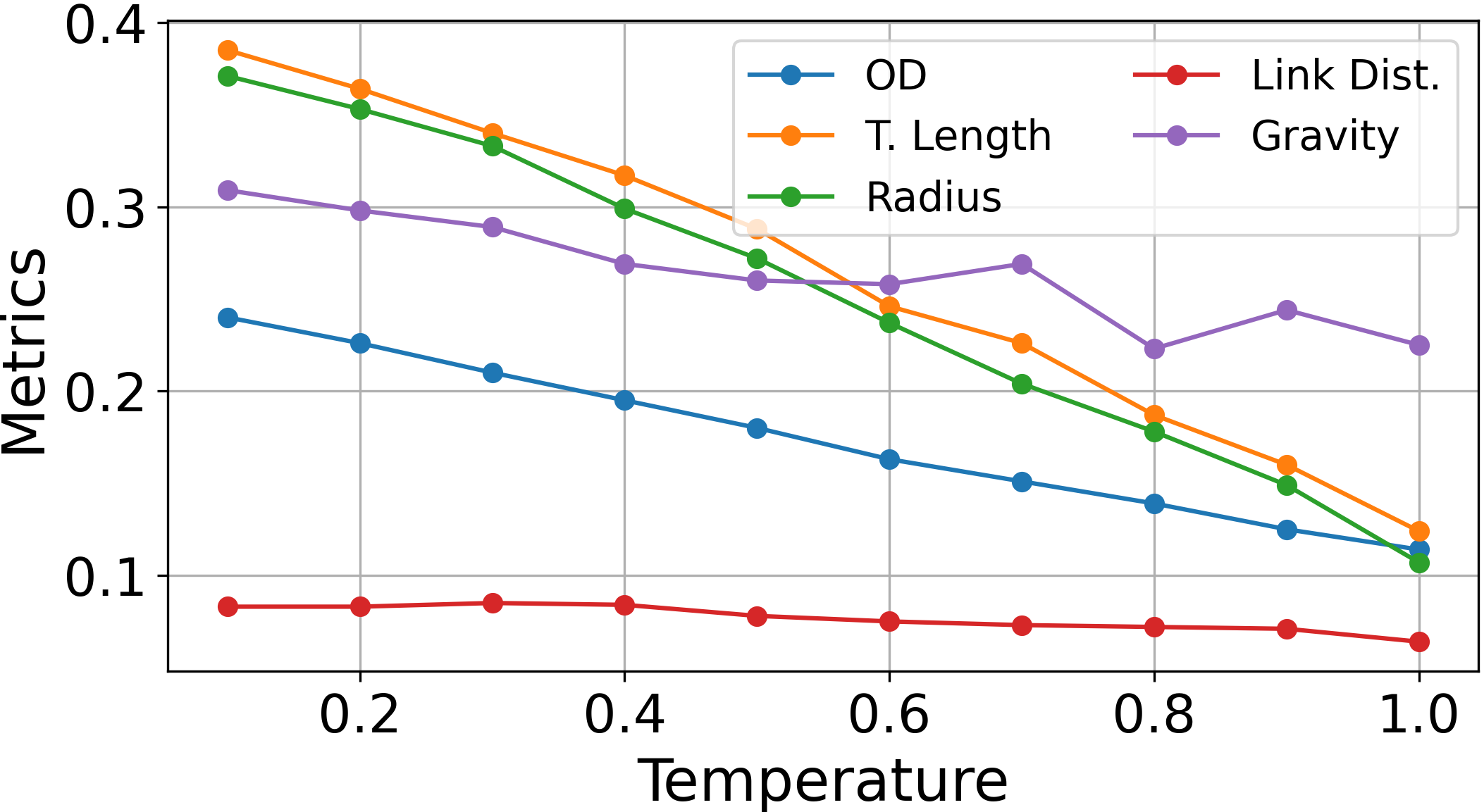}
    \caption{Performance of MobilityGPT under different temperature settings. All metrics use the Jensen-Shannon distributional divergence statistic.}
    \label{fig:temperature}
\end{figure}

Interestingly, as illustrated in Figure \ref{fig:temperature}, MobilityGPT demonstrates a clear trend: its performance tends to increase with higher temperature settings. This observation suggests that mobility data generation can benefit from the randomness of GPT-like models. 

\subsection{Tokenizer Modeling}

\textcolor{black}{In the development of the MobilityGPT tokenizer model, we explored the use of specific tokens to mark the start (\texttt{<BOT>}) and end (\texttt{<EOT>}) of trajectories. Our goal was to assess how these tokens affect the model’s performance in human mobility modeling, particularly for sequential trajectory data. Our findings reveal a key insight: using a single token to indicate a trajectory boundary—whether \texttt{<BOT>} or \texttt{<EOT>}—significantly improves performance compared to using separate start and end tokens. This result indicates that emphasizing trajectory boundaries with a single flag is more beneficial than marking both beginning and end with separate tokens for our mobility modeling task. The effectiveness of a single-token approach highlights the importance of task-specific tokenization strategies in sequential data generation with transformer models. An ablation study, detailed in Table~\ref{tab:ablation_study_token}, compares single- and dual-token approaches using various metrics. This study demonstrates that a single-token strategy outperforms the dual-token method in the MobilityGPT workflow, underscoring the advantages of customized and simplified tokenization.}
\label{app:eot}

\begin{table}[ht]
\centering
\ra{1.2}
\caption{Ablation study on the effect of single-token vs dual-token strategies. All metrics use the Jensen-Shannon divergence.}
\label{tab:ablation_study_token}
\vskip 0.1in
\begin{tabular}{cc>{\columncolor{lightgray}}c>{\columncolor{lightgray}}c>{\columncolor{lightgray}}c>{\columncolor{lightgray}}c>{\columncolor{lightgray}}c}
\Xhline{1pt} % Thick line at the top
\hline
\hline
\texttt{<BOT>} & \texttt{<EOT>} & OD & Trip Length & Radius  & Gravity \\ 
\hline
\textcolor{darkgreen}{\cmark} & \textcolor{darkgreen}{\cmark} & \textbf{0.110} & 0.151 & 0.128 & 0.235 \\
\textcolor{red}{\xmark}  & \textcolor{darkgreen}{\cmark} & 0.114 & \textbf{0.124} & \textbf{0.107} & \textbf{0.225} \\

\hline
\hline
\Xhline{1pt}
\end{tabular}
\end{table}

\subsection{Computational Demand}
\label{s:computations}

\textcolor{black}{Evaluating computational demand is essential to ensure the practical feasibility, scalability, and environmental sustainability of generative models. To ensure fairness, all benchmark models and MobilityGPT were trained on the same hardware setup, NVIDIA TITAN RTX GPU with 24GB memory. MobilityGPT achieves an effective balance between model complexity and efficiency—it is more compact than some of the larger transformer-based models while being more sophisticated than lightweight models like LSTM-based approaches. Regarding training time, MobilityGPT is significantly faster than the more complex baselines, DiffTraj and TS-TrajGen, yet it maintains far superior performance compared to the simplest models.}

\begin{table}[ht]
\centering
\ra{1.3}
\caption{Computational performance of different models vs 
MobilityGPT}

\label{tab:comp_demand}
\vskip 0.1in
\begin{tabular}{@{}lccccc@{}}
\Xhline{1pt} % Thick line at the top
\hline
\hline
& \multicolumn{2}{c}{N\_param} & \multicolumn{2}{c}{Total running time} \\
\cmidrule{2-3} \cmidrule{4-5}
Model & BJ & Porto & BJ & Porto \\
\cmidrule{1-1} \cmidrule{2-3} \cmidrule{4-5}
SeqGAN & 2.6M & 800 & $\sim$ 12hr & $\sim$ 5hr \\
LSTM-TrajGAN & 100k & 100k & $\sim$ 30mins & $\sim$ 1hr \\
TS-TrajGen & 23.2M & 7.6M & $>$ 200hr & $>$ 200hr \\
DiffTraj & 15.7M & 15.7M & $\sim$ 48hr & $\sim$ 60hr \\
\textbf{MobilityGPT} & 5.5M & 5.5M & $\sim$ 5hr & $\sim$ 5hr \\
\hline
\hline
\Xhline{1pt}
\end{tabular}
\end{table}

\textcolor{black}{Overall, MobilityGPT strikes a strong balance between model capacity and computational efficiency. It achieves state-of-the-art performance without the heavy resource requirements of larger models and avoids the accuracy trade-offs seen in lightweight alternatives. This balance makes MobilityGPT a practical and scalable solution for real-world applications where both predictive performance and computational feasibility are critical.}

\section{Discussion}
\label{s:discuss}

\textcolor{black}{While MobilityGPT achieves state-of-the-art results in generating realistic trajectories that respect road network constraints, there are notable limitations. First, the current model does not provide formal privacy guarantees for the synthetic trajectories, which limits its immediate applicability in privacy-sensitive domains such as ride-sharing or urban planning. Addressing this will require integrating differential privacy mechanisms or other formal privacy-preserving techniques into the generative process.}

\textcolor{black}{Second, MobilityGPT generalizes well across diverse urban environments (e.g., Porto and Beijing) when trained and evaluated within the same city. However, its ability to transfer across regions remains limited. A core reason for this is the region-specific nature of the tokenizer, where tokens correspond to individual road segments. Since urban areas vary significantly in road network topology and density, a tokenizer trained on one city cannot be directly applied to another. This restricts the model’s out-of-the-box generalizability and highlights the need for future work on region-agnostic tokenization or transferable representations across cities.}

\textcolor{black}{Lastly, while our current RLTF strategy with trajectory preferences derived from a similarity-based baseline enables scalable and automated optimization, it does not incorporate direct human feedback. Although human-in-the-loop signals could enhance adaptability to evolving mobility trends, collecting such feedback at scale poses practical challenges. We consider this a promising direction for future research, particularly for applications where user alignment and behavioral nuance are critical.}

\section{Conclusion}
\label{s:conclusion}

This paper presents a GPT-based generative method for modeling human mobility characteristics. The proposed mechanism considers a multi-objective learning method with pre-training and fine-tuning stages. MobilityGPT pre-training leverages the understanding of human mobility using the strong self-attention structures by conditioning the trajectory sampling with the gravity model and the sequence sampling with the RCM. To further enhance the MobilityGPT capabilities for generating similar trajectories to real trajectories, we propose a fine-tuning strategy that does not require a human assessment. The novel fine-tuning method trains a reward using the trajectory trip length and employs the reward model as an input to the RL policy gradient method. The experiments validate the effectiveness of the proposed MobilityGPT approach for generating realistic mobility trajectories from real trajectory samples by comparing with state-of-the-art benchmark models.  

Future directions include enhancing the model’s cross-regional generalization capabilities, incorporating robust privacy guarantees and fine-tuning strategies with human feedback, enabling broader applicability in real-world deployments.
\bibliography{references}
\bibliographystyle{IEEEtran}

\begin{IEEEbiography}
[{\includegraphics[width=1in,height=1.4in,clip,keepaspectratio]{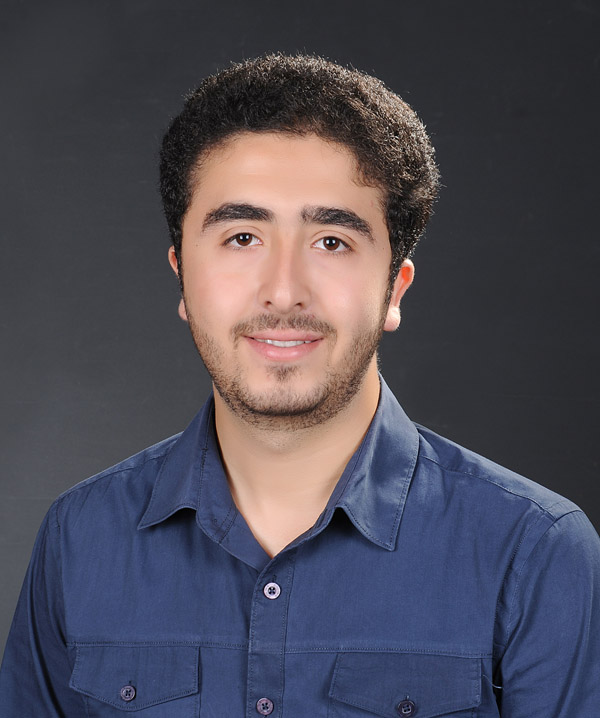}}]{Ammar Haydari}
Received the B.Sc. degree in Electronic Engineering from Uludag University, Bursa, Turkey, in 2014, M.S. degree in Electrical Engineering from the University of South Florida, Tampa, FL, in 2019, and Ph.D. degree from the Department of Electrical and Computer Engineering at the University of California, Davis. His research interests include intelligent transportation systems, deep reinforcement learning, cybersecurity, and machine learning. 
\end{IEEEbiography}

\begin{IEEEbiography}
[{\includegraphics[width=1in,height=1.4in,clip,keepaspectratio]{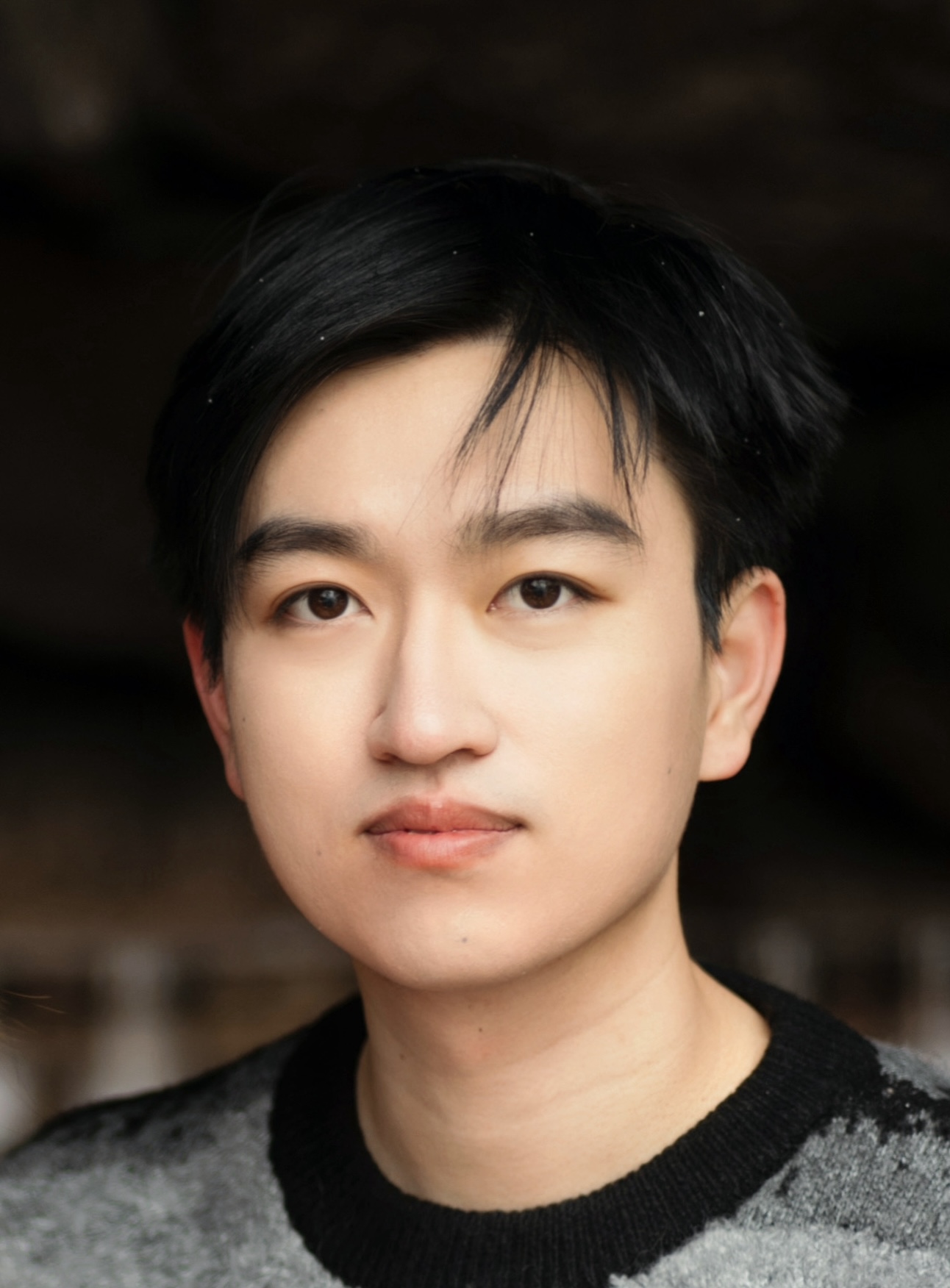}}] {Dongjie Chen} received his Ph.D. degree in Electrical and Computer Engineering from the University of California, Davis. His research focuses on Deep Generative Models, specifically Generative Adversarial Networks (GANs), Large Language Models (LLMs), and Vision-Language Models (VLMs). He also explores the applications of these models in various domains, particularly in healthcare and transportation. 

\end{IEEEbiography}

\begin{IEEEbiography}
[{\includegraphics[width=1in,height=1.4in,clip,keepaspectratio]{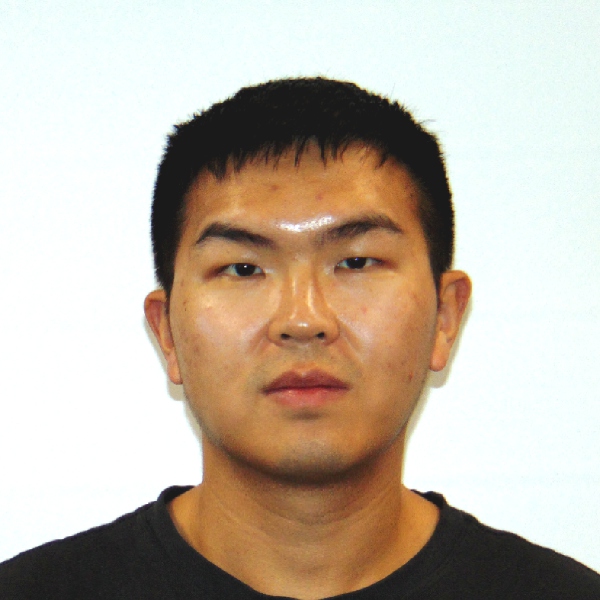}}] {Zhengfeng Lai} received his B.S. in Information Engineering from Zhejiang University in 2019, and his M.S. degree in Electrical and Computer Engineering from the University of California, Davis in 2021. He is currently pursuing his Ph.D. degree at the University of California, Davis. He has served as a reviewer for the IEEE Trans. of Image Processing, ICML, ICLR, NeurIPS, ICCV, CVPR, and ECCV. 
\end{IEEEbiography}

\begin{IEEEbiography}
[{\includegraphics[width=1in,height=1.4in,clip,keepaspectratio]{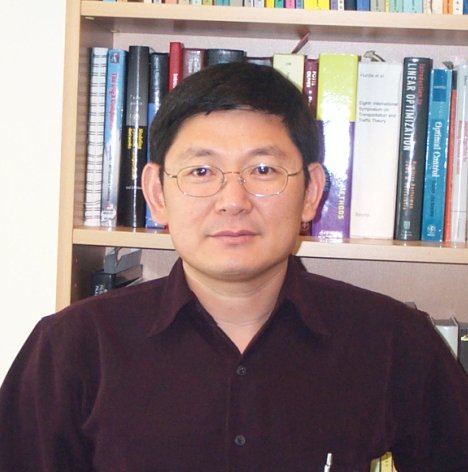}}]{Michael Zhang}is currently a professor in the Civil and Environmental Engineering Department at University of California Davis. His research is in traffic operations and control, transportation network analysis and intelligent transportation systems. Professor Zhang received his BS degree in Civil Engineering from Tongji University, and MS and PhD degrees in Engineering from University of California Irvine. He is an Area Editor of the journal Network and Spatial Economics, and an Associate Editor of Transportation Research, Part B, Methodological, and Transportation Science.
\end{IEEEbiography}

\begin{IEEEbiography}
[{\includegraphics[width=1in,height=1.4in,clip,keepaspectratio]{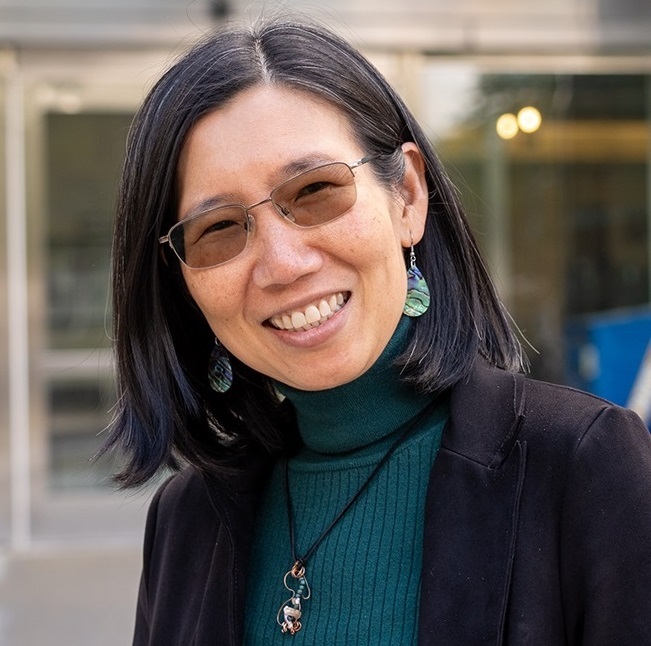}}] {Chen-Nee Chuah} received her B.S. in Electrical Engineering from Rutgers University, and her M. S. and Ph.D. in Electrical Engineering and Computer Sciences from the University of California, Berkeley. She is currently the Child Family Professor in Engineering in Electrical and Computer Engineering at the University of California, Davis. Her research interests lie in applying data analytics and AI/ML techniques to complex networked systems and applications, including cyberinfrastructure, online social platforms, healthcare, and intelligent transportation systems. Chuah is a Fellow of the AAAS and IEEE, and an ACM Distinguished Scientist. She currently serves as an Associate Editor for the ACM Transactions on Computing for Healthcare and has previously served on the editorial board for IEEE/ACM Transactions on Networking, IEEE Transactions on Mobile Computing, and IEEE Internet of Things Journal.
\end{IEEEbiography}

\end{document}